\documentclass[letterpaper]{IEEEtran}
\usepackage{multirow}
\usepackage{graphicx}
\usepackage{amsmath}
\usepackage{amssymb}
\usepackage{booktabs}
\usepackage{pifont}
\usepackage{textcomp}
\usepackage{tabularx}
\usepackage{tikz}
\usepackage{makecell}
\usepackage{pgfplots}
\pgfplotsset{compat = 1.18,
    tick label style = {font=\scriptsize},
    every axis label = {font=\scriptsize},
    legend style = {font=\scriptsize},
    label style = {font=\scriptsize}
}
\usepackage{float}

\usepackage[pagebackref,breaklinks,colorlinks]{hyperref}
\usetikzlibrary{arrows.meta}

\usepackage[capitalize]{cleveref}
\usepackage{caption}
\usepackage{subcaption}
\crefname{section}{Sec.}{Secs.}
\Crefname{section}{Section}{Sections}
\Crefname{table}{Table}{Tables}
\crefname{table}{Tab.}{Tabs.}
\usepackage{flushend}
\usepackage{color}

\begin{document}

\title{TruckV2X: A Truck-Centered Perception Dataset}

\author{Tenghui Xie$^{1}$,~Zhiying Song$^{1}$,~Fuxi Wen$^{1}$,~\IEEEmembership{Senior~Member,~IEEE},~Jun Li$^{1}$,~Guangzhao Liu$^{2}$ and Zijian Zhao$^{2}$
\thanks{
Manuscript received: March 18, 2025; Revised: May 23, 2025; Accepted: July 11, 2025. This paper was recommended for publication by Editor Hyungpil Moon upon evaluation of the Associate Editor and Reviewers' comments. \textit{(Corresponding author: Fuxi Wen)}
}
\thanks{$^{1}$Tenghui Xie, Zhiying Song, Fuxi Wen, and Jun Li are with the School of Vehicle and Mobility, Tsinghua University, Beijing 100190, China (e-mail: wenfuxi@tsinghua.edu.cn).}
\thanks{$^{2}$Guangzhao Liu and Zijian Zhao are with FAW Jiefang Group Co., Ltd., Changchun 130011, China.}
\thanks{Digital Object Identifier (DOI): see top of this page.}
}
\markboth{IEEE Robotics and Automation Letters. Preprint Version. July, 2025}
{Xie \MakeLowercase{\textit{et al.}}: TruckV2X: A Truck-Centered Perception Dataset} 
\maketitle

\begin{abstract}
Autonomous trucking offers significant benefits, such as improved safety and reduced costs, but faces unique perception challenges due to trucks' large size and dynamic trailer movements. These challenges include extensive blind spots and occlusions that hinder the truck's perception and the capabilities of other road users. To address these limitations, cooperative perception emerges as a promising solution. However, existing datasets predominantly feature light vehicle interactions or lack multi-agent configurations for heavy-duty vehicle scenarios.
To bridge this gap, we introduce TruckV2X, the first large-scale truck-centered cooperative perception dataset featuring multi-modal sensing (LiDAR and cameras) and multi-agent cooperation (tractors, trailers, CAVs, and RSUs). We further investigate how trucks influence collaborative perception needs, establishing performance benchmarks while suggesting research priorities for heavy vehicle perception. The dataset provides a foundation for developing cooperative perception systems with enhanced occlusion handling capabilities, and accelerates the deployment of multi-agent autonomous trucking systems.
The TruckV2X dataset is available at \href{https://huggingface.co/datasets/XieTenghu1/TruckV2X}{https://huggingface.co/datasets/XieTenghu1/TruckV2X}.
\end{abstract}


\section{Introduction}
\IEEEPARstart{A}{utonomous} trucking is expected to benefit the logistics industry in improved road safety, reduced operational costs, and solutions to driver shortages \cite{slowik2018automation}. As an indispensable yet high-impact participant in modern transportation systems, the safe deployment of intelligent trucks hinges on two critical factors: {their ability to comprehensively perceive their surroundings and to operate without impairing the safety or perceptual abilities of other road users.}

\begin{figure}[t]
    \centering
    \subfloat[Trailer obstructs tractor.]{
    \includegraphics{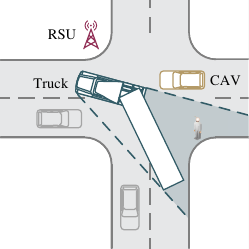}
    \label{fig:1a}}
    \subfloat[Truck obstructs others.]{
    \includegraphics{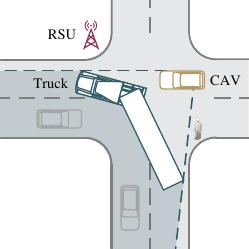}
    \label{fig:1b}}
    \vspace{2mm}
    \subfloat[Occluded area at different positions within a circle of radius $R$.]{
    \includegraphics[width=0.98\linewidth]{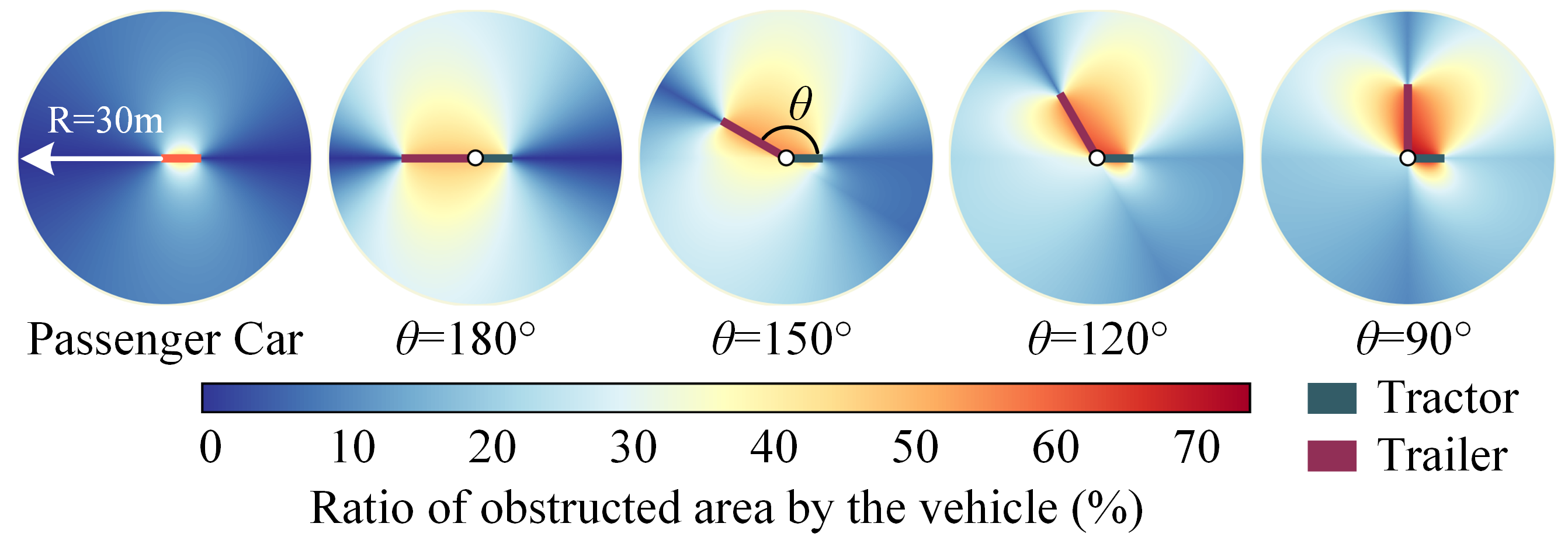}
    \label{fig:theo_occ}
    }
    \caption{Illustration of truck-related occlusions.}
    \label{fig:obstract_ill}
\end{figure}

These factors emerge from the operational constraints of tractor-trailer systems. Their substantial dimensions create persistent blind zones (Figure \ref{fig:1a}, \ref{fig:1b}), where structural components obstruct both self-perception and neighboring agents' sensing capabilities. Articulated motion further exacerbates occlusions through dynamic spatial conflicts between tractors and trailers during maneuvers \cite{ehlgen2008eliminating}. Compounding these issues, low-speed operation combined with large spatial footprints induces sustained environmental uncertainties in mixed traffic, systematically expanding risk zones \cite{ahrholdt2009intersection}.

As an example, Figure \ref{fig:theo_occ} illustrates how visibility is obstructed within a 30-meter observational range. While standard passenger cars retain a relatively clear view ($<$30\% occlusion area), tractor-trailers display 1.5$\times$ greater baseline occlusion even when driving straight. The problem intensifies as trailers pivot: angular displacement between tractor and trailer steeply degrades visibility, with occlusion surpassing 70\% during 90$^\circ$ turns. These dual challenges, sheer vehicle size, and pivoting motion patterns, collectively position articulated trucks as high-risk urban blind zone generators.

Numerous efforts have been made in human-driven trucks to design blind-spot detection systems \cite{BoschMobility2023,ZF2023}. There are two potential directions to further enhance the perception intelligence of trucks. First, advanced data-driven algorithms are needed to achieve comprehensive perception by integrating information from the tractor and trailer. Second, cooperative perception systems can enable trucks to share data with other road users and infrastructures. By leveraging Vehicle-to-Everything (V2X) communication technology, we can connect the tractor and trailer internally, allowing them to perceive the environment collaboratively. Additionally, trucks can communicate with external agents, such as connected and automated vehicles (CAVs) and roadside units (RSUs), enabling seamless sensor data integration from diverse sources. This integration of multi-perspective information broadens the perceptual range of all agents and helps overcome occlusions, enhancing overall traffic safety and efficiency.

\begin{figure}[t]
    \centering
    \begin{tikzpicture}[
    every node/.style={font=\footnotesize},
    discript/.style 2 args={ 
        rectangle,
        color=black,
        minimum size=6pt,
        fill=none,
        inner sep=0,
        label={[text=black,align=center, inner xsep=7pt, text height=13pt, anchor=west]#2}
    },
    single/.style 2 args={ 
        rectangle,
        fill=#1,  
        color=gray!80,
        minimum size=6pt,
        inner sep=0,
        label={[text=black,align=center, inner xsep=7pt, text height=13pt, anchor=west]#2}
    },
    v2x/.style 2 args={ 
        circle,
        fill=#1,
        color=red!80!black,
        minimum size=6pt,
        inner sep=0,
        label={[text=black,align=center, inner xsep=7pt, text height=13pt, anchor=west]#2}
   },
    v2v/.style 2 args={ 
        rectangle,
        draw=#1,  
        color=purple!80!black,
        thick,
        minimum size=6pt,
        inner sep=0,
        fill=none, 
        label={[text=black,align=center, inner xsep=7pt, text height=13pt, anchor=west]#2}
    },
    v2i/.style 2 args={ 
        circle,
        draw=#1,  
        color=blue!80!black,
        thick,
        minimum size=6pt,
        inner sep=0,
        fill=none, 
        label={[text=black,align=center, inner xsep=7pt, text height=13pt, anchor=west]#2}
    }
]
    \draw[latex-latex,thick] (-4.2,0) -- (4.2,0);
    \draw[latex-latex,thick] (0,-2.1) -- (0,2.2);
    
    \node[anchor=south] at (3.2,0) {\textbf{Heavy Vehicle}};
    \node[anchor=south] at (-3.2,0) {\textbf{Light Vehicle}};
    \node[anchor=south] at (0,2.2) {\textbf{Multi-Agent}};
    \node[anchor=north] at (0,-2.1) {\textbf{Single-Agent}};

    \node[v2x={}{\underline{{TruckV2X}}}] at (0.9,1.2) {};

    \node[v2v={}{OPV2V \cite{xu2022opv2v}}] at (-3,1.7) {};
    \node[v2i={}{DAIR-V2X \cite{yu2022dair}}] at (-3,1.2) {};
    \node[v2x={}{V2X-Real \cite{xiang2025v2x}}] at (-3,0.7) {};

    \node[single={}{KITTI \cite{geiger2013vision}}] at (-3,-0.4) {};
    \node[single={}{nuScenes \cite{caesar2020nuscenes}}] at (-3,-0.9) {};
    \node[single={}{Waymo \cite{sun2020scalability}}] at (-3,-1.4) {};

    \node[single={violet!80!black}{MAN TruckScenes \cite{fent2024man}}] at (0.5,-0.9) {};

    \node[single={}{\textit{Single vehicle}}] at (-3.2,-2.8) {};
    \node[v2v={}{\textit{V2V}}] at (-0.8,-2.8) {};
    \node[v2i={}{\textit{V2I}}] at (0.6,-2.8) {};
    \node[v2x={}{\textit{V2V and V2I}}] at (1.8,-2.8) {};

\end{tikzpicture}
    \caption{Datasets available for the perception of autonomous driving. TruckV2X is the first multi-agent collaborative perception dataset specifically designed for heavy vehicles.  (V2V: Vehicle-to-Vehicle, V2I: Vehicle-to-Infrastructure) }
    \label{fig:dataset_comp}
\end{figure}
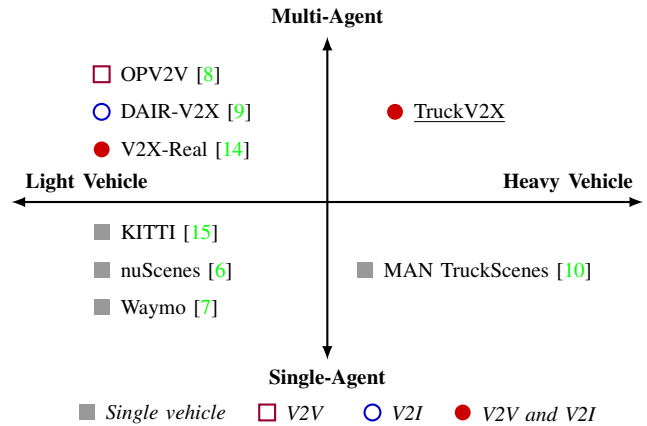

However, the development of such systems is significantly impeded by a lack of suitable datasets. As highlighted in Figure \ref{fig:dataset_comp}, existing datasets, such as nuScenes \cite{caesar2020nuscenes} and Waymo Open Dataset \cite{sun2020scalability}, are primarily designed for light vehicles and fail to account for the unique challenges of trucks. Similarly, cooperative perception datasets like OPV2V \cite{xu2022opv2v} and DAIR-V2X \cite{yu2022dair} focus on light vehicles, while truck-specific datasets such as Man TruckScenes \cite{fent2024man} are limited to single-agent setups with sensors mounted only on the tractor, ignoring the critical role of trailers and the need for cooperative perception. 

To bridge these gaps, we present TruckV2X, the first truck-centered perception dataset, featuring multi-modal sensing (LiDAR and cameras) and multi-agent cooperation (tractors, trailers, CAVs, and RSUs). 
Utilizing Unreal Engine, we modeled a semi-trailer truck and integrated it into the CARLA simulation environment \cite{dosovitskiy2017carla} as an interactive agent. The truck has two 64-channel LiDARs and five surround-view cameras on both the tractor and the trailer. Other agents, including one CAV and one RSU, are also equipped with LiDARs and multi-view cameras.
The dataset comprises 64 scenarios, featuring 88,396 frames of LiDAR point clouds, one million camera images, and 1.18 million 3D bounding box annotations. Each object is annotated with a bounding box, ID, speed, and other relevant information, supporting tasks such as object detection and tracking. Additionally, we provide eight benchmark implementations for cooperative perception tasks and conduct extensive experimental analysis on occlusion scenarios, demonstrating the critical value of truck-specific viewpoints for cooperative perception.

Overall, the main contributions of this work include: 
\begin{itemize}
\item We propose TruckV2X, the first truck-centered multi-modal and multi-agent cooperative perception dataset, addressing data scarcity for truck-specific challenges such as trailer occlusions and multi-agent cooperation.
\item We provide a benchmark for truck-centered collaborative perception to encourage further research and advance cooperative technologies for truck perception safety.
\item Through quantitative analysis, we propose the concept of trucks as mobile perception platforms, demonstrating their dual role as both occlusion sources and enhancers in reducing environmental occlusions.
\end{itemize}

\section{Related Work}
\emph{Perception of truck:}
Previous studies have developed perception-assisted systems in human-driven trucks to reduce blind spots and enhance state estimation capabilities. These systems often rely on additional sensors, such as millimeter-wave radar and cameras, to improve driver awareness and safety \cite{BoschMobility2023, ZF2023}. Recent advancements have also explored the integration of computer vision for more robust perception and state estimation \cite{dong2024dsvt}.
In terms of datasets, despite progress in autonomous driving perception, most datasets focus on light vehicles, leaving a significant gap. TuSimple \cite{yoo2020end} includes some heavy vehicle data but is limited in scope and volume. MAN TruckScenes \cite{fent2024man}, the first large-scale multi-modal dataset for heavy vehicles, lacks comprehensive trailer-mounted sensor data and multi-agent setups. In this work, TruckV2X provides multi-modal data from the tractors, trailers, CAVs, and RSUs, offering a foundation for blind spot detection, state estimation, and environmental perception in heavy vehicles.

\emph{V2X perception and datasets:} V2X technology has been applied in heavy vehicles for tasks such as collision warning \cite{shooter2009intersafe} and platooning \cite{braiteh2024platooning}. While V2X systems primarily transmit basic information, no studies have proposed datasets for high-level cooperative perception that include trucks as cooperative agents \cite{song2025traf}.
On the other hand, recent years have seen significant progress in cooperative perception datasets for passenger cars: OPV2V \cite{xu2022opv2v} pioneered synthesized multi-vehicle perception data, while DAIR-V2X \cite{yu2022dair} marked the first real-world V2I dataset for object detection. Numerous other datasets have since been introduced \cite{zimmer2024tumtraf, xiang2025v2x,yang2024v2x}. They focus on light vehicles, leaving a critical gap for heavy vehicles.
In cooperative perception, heavy vehicles present unique challenges, such as significant occlusions caused by the large size and dynamic movements of tractors and trailers. These occlusions not only hinder the truck's perception but also disrupt other agents in the cooperative system.
The proposed dataset, TruckV2X, bridges these gaps by generating large-scale occlusion scenarios and providing multi-view data from tractors, trailers, CAVs, and RSUs, offering a foundation for advancing cooperative perception in heavy vehicles.

\section{Dataset}
\subsection{Vehicle configurations}

TruckV2X is the first to incorporate semi-trailer trucks into autonomous driving simulation environments for cooperative perception applications. Focusing on urban-prevalent container trucks (tractor-trailer combinations), we developed high-fidelity models through a sequential workflow. First, using Blender, we constructed and optimized articulated vehicle skeletons with reduced mesh complexity. These models were then imported into Unreal Engine, where the tractor and trailer were linked via physics-based kinematic constraints within the Blueprint system. Concurrently, we configured visual properties (customizable colors, materials, and lighting responses) and dynamic parameters (mass distribution, suspension, and powertrain characteristics) to align with real-world truck specifications. The key parameters are documented in Table \ref{tab:vehicle_setup}. In this paper, we regard tractor and trailer as separate agents considering real-world operational dynamics, where tractors frequently engage with different trailers.

\begin{table}[t]
\footnotesize
\centering
\caption{Key parameters of the truck in our dataset.}
\definecolor{color}{RGB}{255, 0, 0}   
\setlength{\tabcolsep}{1.8pt} 
\renewcommand\arraystretch{1.1} 
\begin{tabular}{c|c|c|c|c|c}
\toprule
     & Weight (kg) &Length (m)&Width (m)& Height (m)&CG Height (m)\\
     \hline     Tractor&8,805&6.51&3.05&4.20&2.18 \\
     Trailer&20,000&14.25&2.71&4.40&2.06\\
     \bottomrule
\end{tabular}

\label{tab:vehicle_setup}
\end{table}

\begin{table}[t]
\footnotesize
\centering
\caption{Sensor specifications for each agent.}
\definecolor{color}{RGB}{255, 0, 0}   
\setlength{\tabcolsep}{2.2pt} 
\renewcommand\arraystretch{1.1} 

\begin{tabular}{c|c|c}
    \toprule
    {\textbf{Agents}} &{\textbf{Sensors}} &{\textbf{Details}}
    \\
    \hline  
    \multirow{5}{*}{\makecell{Tractor\\/\\Trailer}} 
    &\makecell{1$\times$ {Camera} \\ (front/rear)} 
    &\makecell[l]{RGB, Resolution: 800$\times$600, FOV:90\textdegree}
    \\
    & \makecell{4$\times$ {Camera} \\ (side)} 
    &{\makecell[l]{RGB, Resolution: 800$\times$600, FOV:120\textdegree}}
    \\
    & 2$\times$ {LiDAR} 
    & \makecell[l]{64 channels, 10\,$\mathrm{Hz}$ capture frequency, \\ 360$^\circ$ horizontal FOV, $-30^\circ$ to $5^\circ$ vertical FOV, \\ 1.536\,$\mathrm{M}$ points per second, 120\,$\mathrm{m}$ capturing range}
    \\
    \hline
    \multirow{4}{*}{CAV} 
    & 4$\times${Camera} 
    &{\makecell[l]{RGB, Resolution: 800$\times$600, FOV:90\textdegree}}
    \\
    & {1$\times${LiDAR}} 
    & \makecell[l]{64 channels, 10\,$\mathrm{Hz}$ capture frequency, \\ 360$^\circ$ horizontal FOV, $-30^\circ$ to $10^\circ$ vertical FOV, \\ 1.536\,$\mathrm{M}$ points per second, 120\,$\mathrm{m}$ capturing range}
    \\
    \hline
    \multirow{4}{*}{RSU} 
    & 1$\times${Camera} 
    &{\makecell[l]{RGB, Resolution: 800$\times$600, FOV:120\textdegree}}
    \\
    & {1$\times${LiDAR}}
    & \makecell[l]{64 channels, 10\,$\mathrm{Hz}$ capture frequency, \\ 360$^\circ$ horizontal FOV, $-30^\circ$ to $5^\circ$ vertical FOV, \\ 1.536\,$\mathrm{M}$ points per second, 120\,$\mathrm{m}$ capturing range}
    \\
\bottomrule
\end{tabular}
\label{tab:sensors}
\end{table}

\begin{figure}[t]
    \centering
    \includegraphics{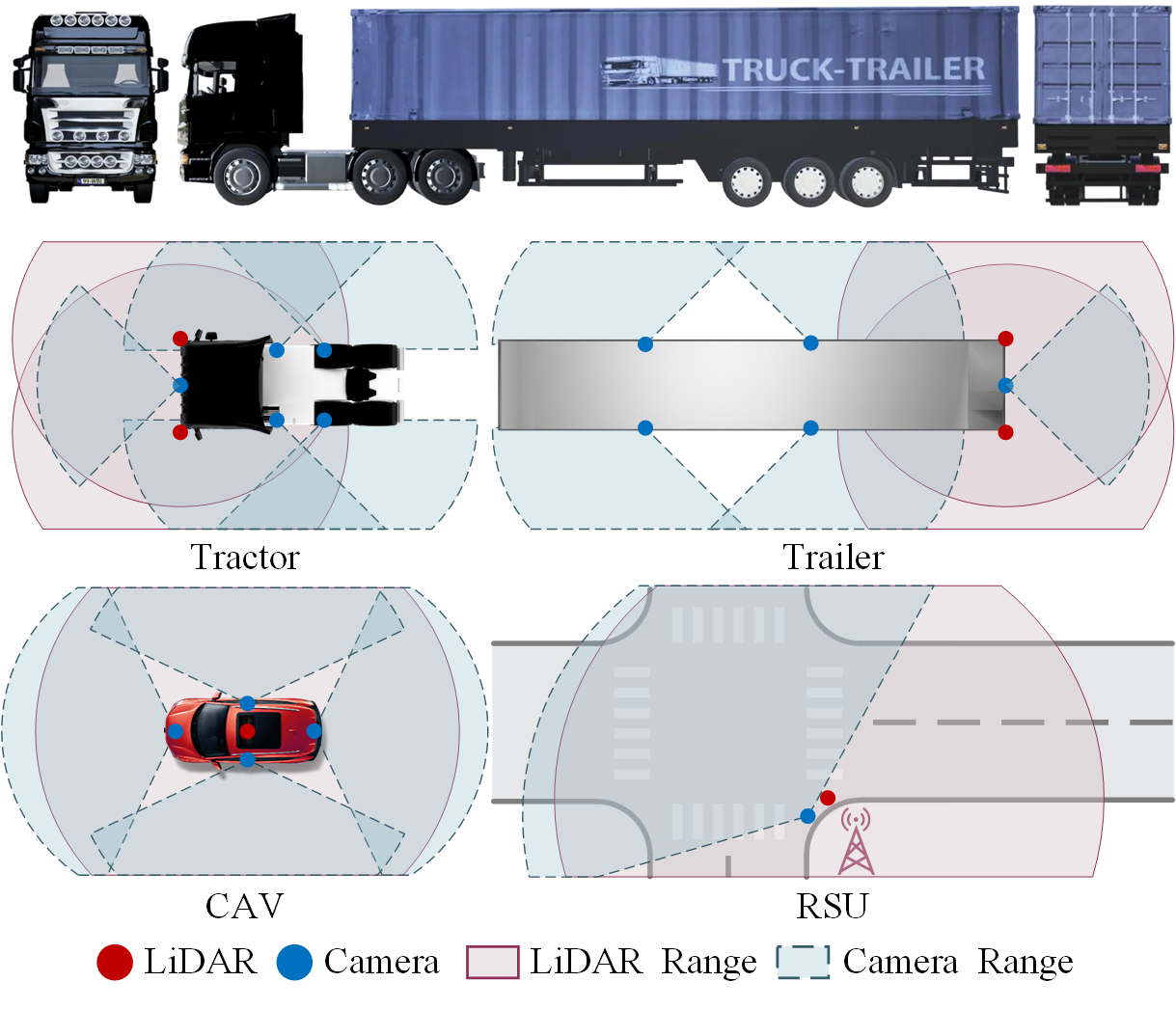}
    \caption{Illustration of sensor configurations for the agents.}
    \label{fig:sensors}
\end{figure}

\subsection{Sensor configurations}

Multi-modal data from multiple perspectives form the cornerstone of cooperative perception systems. To ensure comprehensive and reliable data acquisition, each agent (tractor, trailer, CAV, and RSU) is equipped with a carefully designed sensor suite comprising LiDAR, RGB cameras, and IMU units. Table \ref{tab:sensors} summarizes the detailed sensor specifications, and Figure \ref{fig:sensors} illustrates the sensor coordinate system and spatial layout across all platforms.

For CAVs, the baseline configuration includes a roof-mounted LiDAR paired with four strategically positioned RGB cameras providing full 360° coverage. In contrast, the truck's extended physical dimensions (tractor and trailer) necessitate a distributed sensing architecture: two LiDARs are installed at the upper front corners of the tractor, while two additional LiDARs monitor the trailer's rear quadrant. This configuration is complemented by five RGB cameras – two on each lateral side and one at the tractor's front/trailer's rear – ensuring continuous environmental monitoring across the articulated vehicle's entire length. 
The RSU configuration integrates a LiDAR and a camera oriented toward traffic flow. To optimize coverage and cost-effectiveness, the majority of RSUs in our dataset are deployed on traffic light poles at lane centers or intersection corners. Considering the high pedestrian density in urban scenarios, the LiDAR is mounted at a four-meter height to minimize blind spots in close-range sensing while maintaining practical coverage. This balanced design enhances detection reliability at urban intersections where frequent vehicle-pedestrian interactions occur.

\begin{figure*}[t]
    \raggedright
    \hspace{-5pt}
    \subfloat[T-intersection]{
    \includegraphics[scale=1.27]{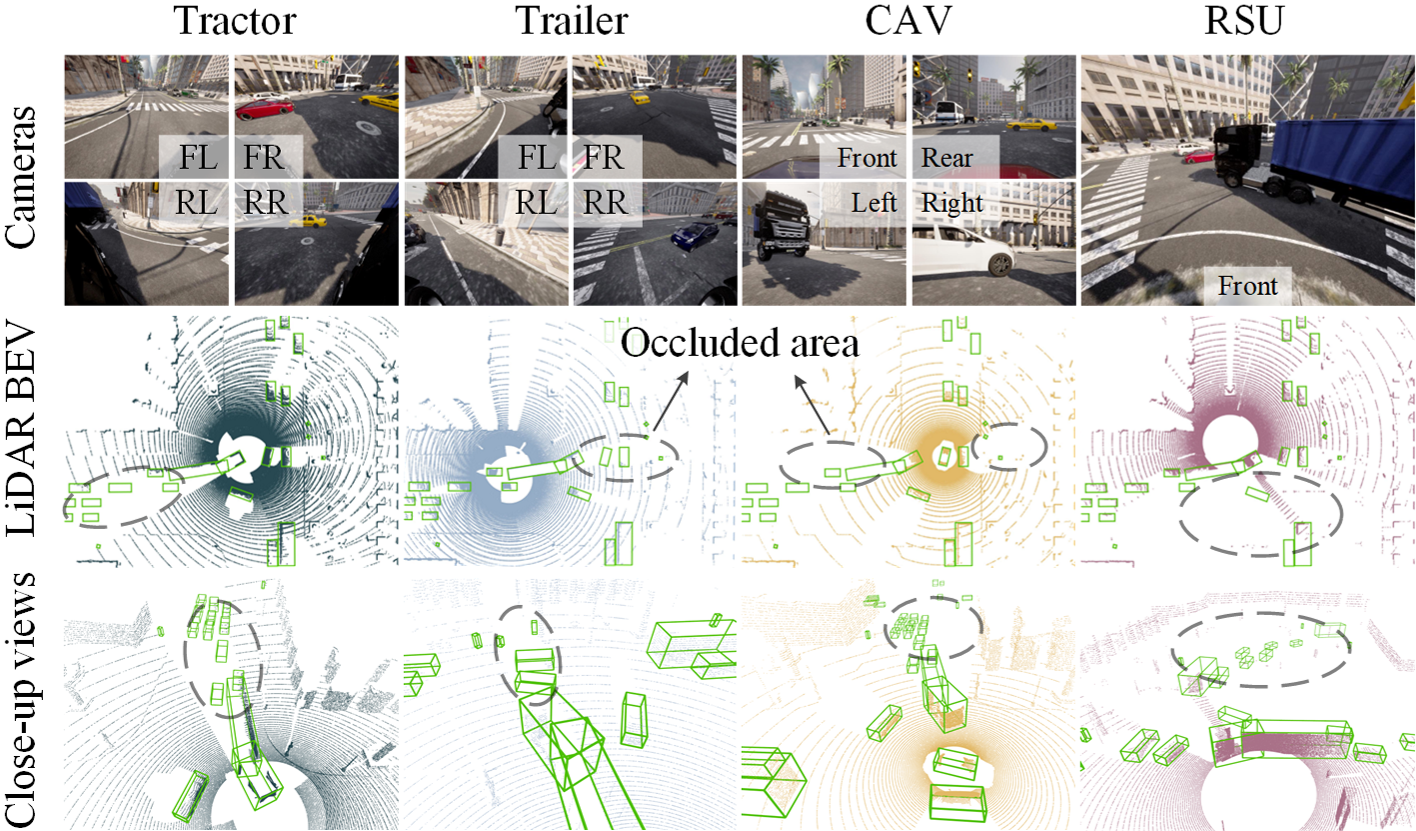}}
    \subfloat[Straight road]{
    \includegraphics[scale=1.27]{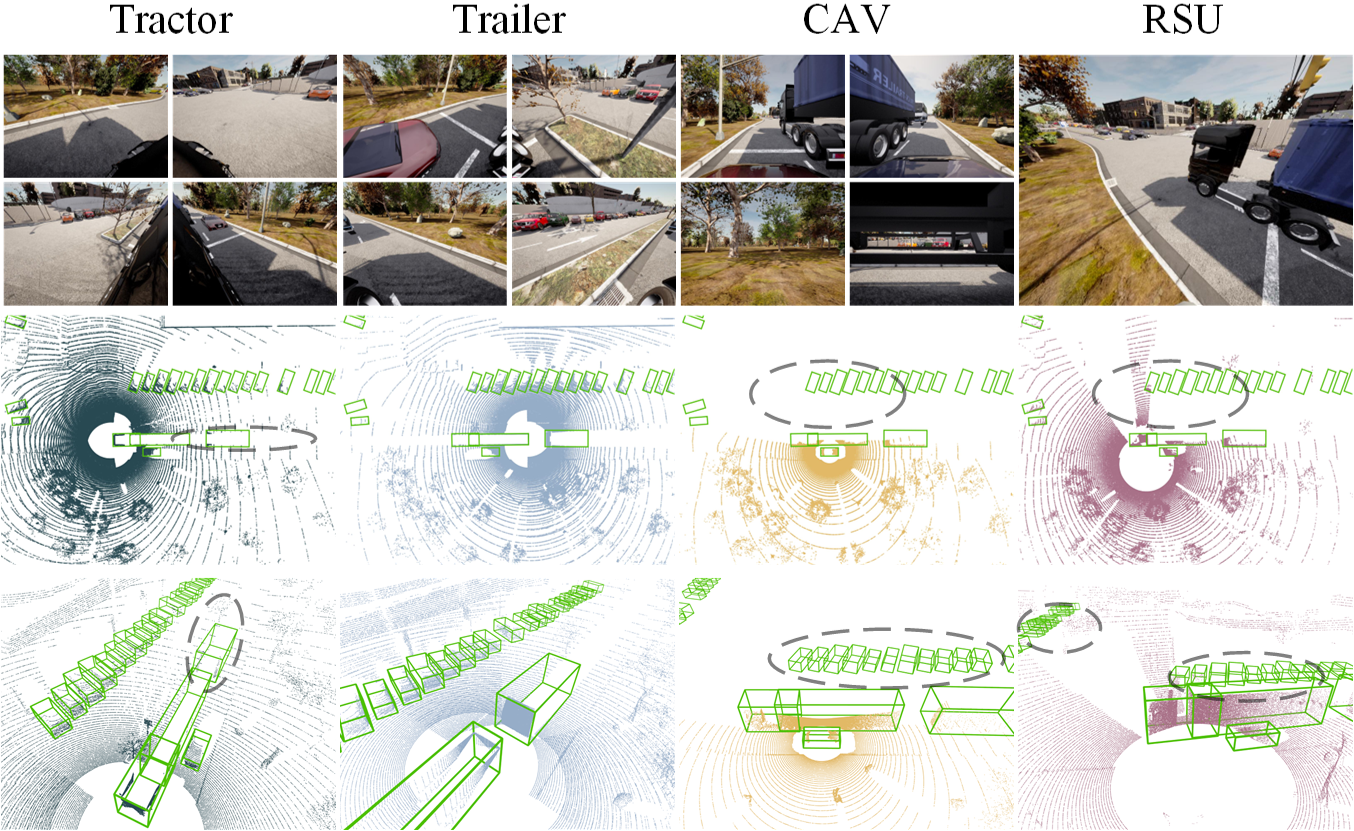}}
    
    \vfill
    \vspace{1mm}
    \hspace{-5pt}
    \subfloat[Four-way intersection]{
    \includegraphics[scale=1.27]{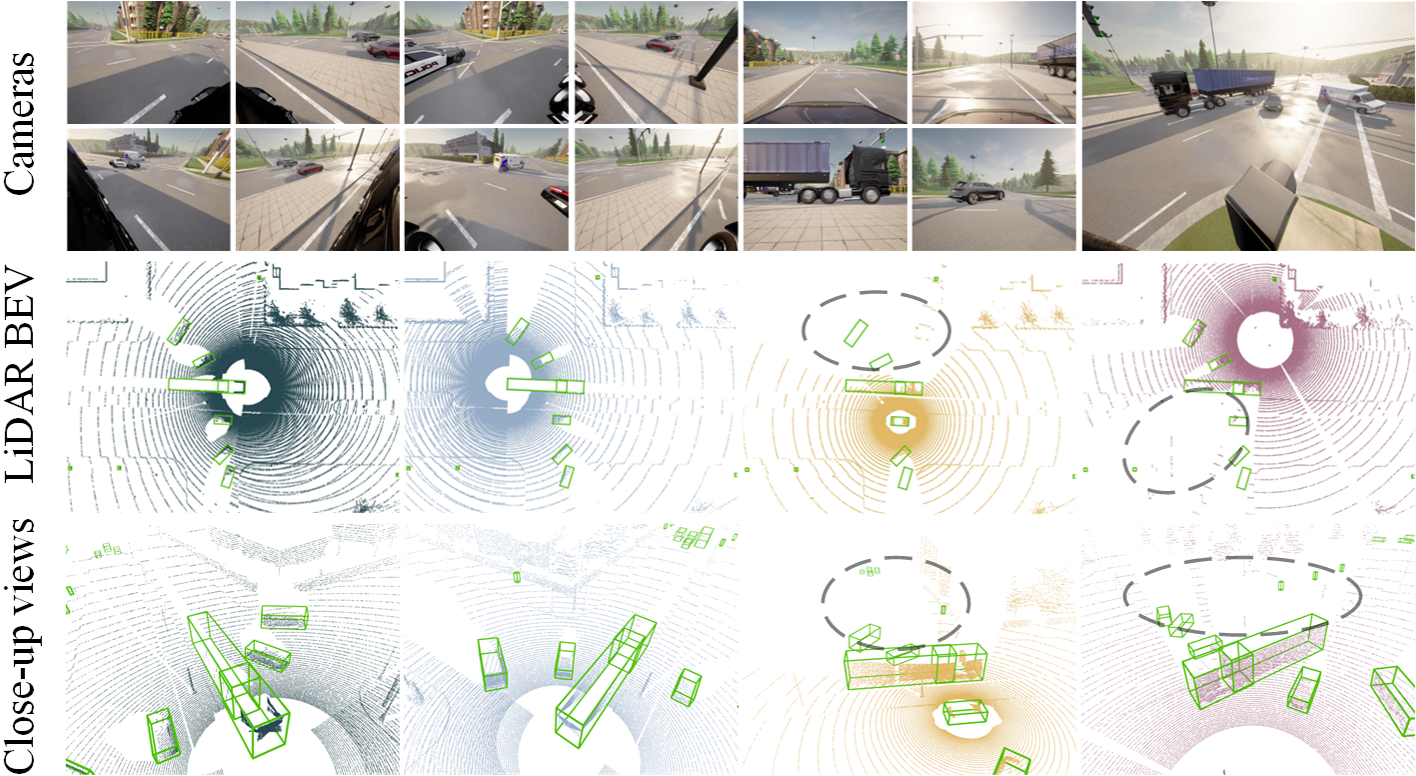}}
    \subfloat[Roundabout]{
    \includegraphics[scale=1.27]{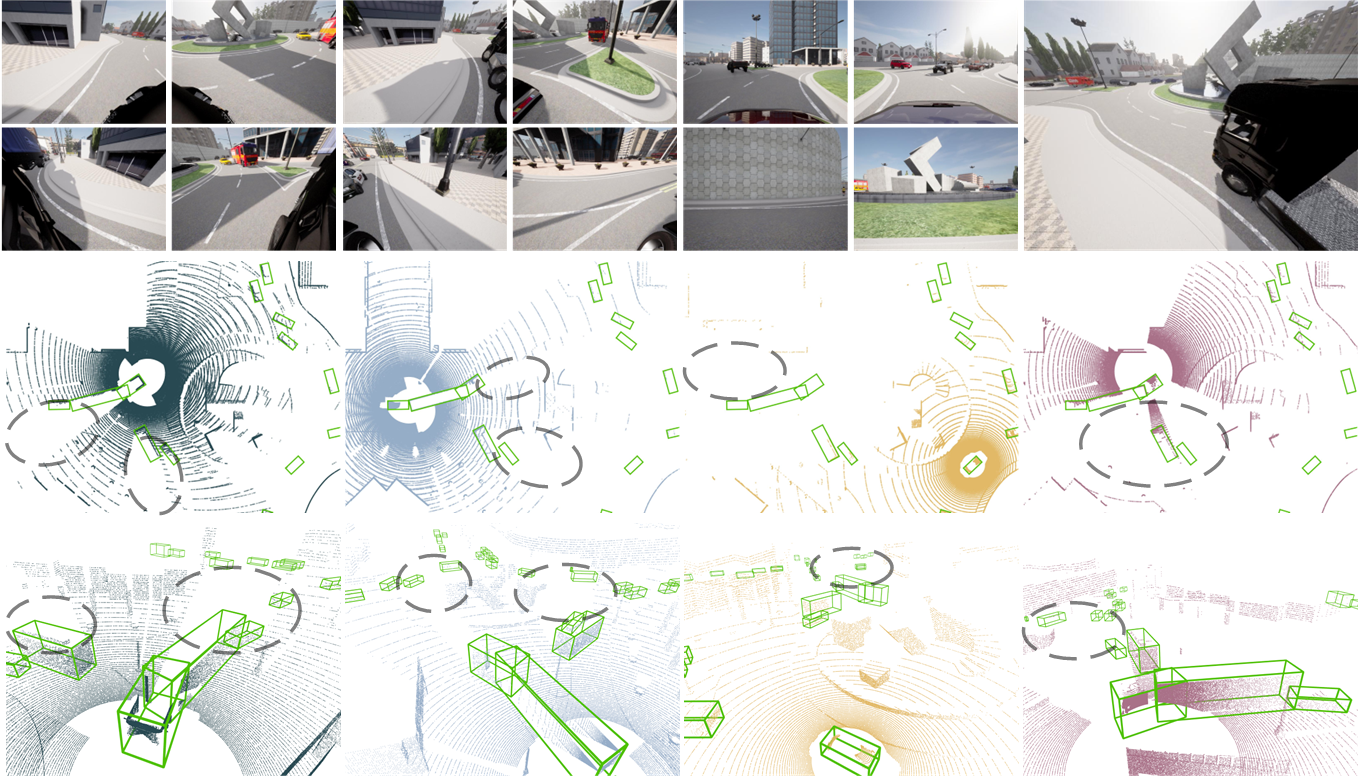}}
    \caption{Four representative scenarios in our dataset: T-intersection, straight road, four-way intersection, and roundabout. In each scenario, the top row shows RGB images from agent perspectives; the middle row displays BEV projections of LiDAR point clouds; the bottom row highlights occluded areas in close-up views, where dashed ellipses annotate critical blind zones for each agent (trailer obstructs tractor and truck obstructs others). All scenarios present the agent sequence from left to right as tractor, trailer, CAV, and RSU.}
    \label{fig:occlusion_scene}
\end{figure*}

\subsection{Scenarios}

\begin{figure*}[t]
    \raggedright
    \hspace{0pt}
    \begin{subfigure}[t]{0.23\textwidth}
        \centering
%
%
%
\definecolor{mycolor1}{rgb}{0.20000,0.36000,0.40000}%
\begin{tikzpicture}[font=\scriptsize]
\begin{axis}[%
major tick length=1.5pt,
width=2.0in,
height=1.2in,
at={(0in,0in)},
xmin=125,
xmax=675,
xlabel style={yshift=0.08in},
xlabel={Frames per scenario},
ymin=0,
ymax=13.5,
ylabel style={yshift=-0.1in},
ylabel={Counts},
axis background/.style={fill=white},
xmajorgrids=false,
ymajorgrids=true,
grid style={dashed, opacity=0.6},
ytick distance=3
]
\addplot[ybar interval, fill=mycolor1, fill opacity=0.5, draw=black, area legend, line width=0.7pt] table[row sep=crcr] {%
x	y\\
150	2\\
200	10\\
250	13\\
300	11\\
350	7\\
400	11\\
450	3\\
500	2\\
550	3\\
600	2\\
650	2\\
};
\end{axis}
\end{tikzpicture}%
        \vspace{-20pt}
        \caption{}
        \label{fig:scene}
    \end{subfigure}
    \hspace{-3pt}
    \begin{subfigure}[t]{0.23\textwidth}
        \centering
%
%
\definecolor{mycolor1}{rgb}{0.70588,0.60392,0.40392}%
\definecolor{mycolor2}{rgb}{0.56863,0.18431,0.33725}%
\definecolor{mycolor3}{rgb}{0.20000,0.36078,0.40392}%
\begin{tikzpicture}[font=\scriptsize]

\begin{axis}[%
major tick length=1.5pt,
width=2.0in,
height=1.2in,
at={(0in,0in)},
xmin=-3,
xmax=200,
xlabel={Distance (m)},
ymin=0,
ymax=7400,
ylabel={Counts},
axis background/.style={fill=white},
xmajorgrids=false,
ymajorgrids=true,
grid style={dashed, opacity=0.6},
legend style={fill=none, nodes={scale=1.0},legend cell align=left, align=left, font=\tiny,inner sep=0pt,outer sep=0pt, row sep=-3pt},
xlabel style={yshift=0.08in},
ylabel style={yshift=-0.08in},
ytick distance=1500,
scaled y ticks=base 10:-3
]
\addplot [color=mycolor3, line width=1.0pt]
  table[row sep=crcr]{%
0	0\\
6.30000000000018	1605\\
18.8999999999996	5274\\
31.5	5234\\
44.1000000000004	3188\\
56.6999999999998	2838\\
69.3000000000002	1180\\
81.8999999999996	605\\
94.5	319\\
107.1	264\\
119.7	142\\
132.3	67\\
144.9	98\\
157.5	415\\
170.1	311\\
182.7	186\\
195.3	118\\
207.9	91\\
220.5	83\\
233.1	58\\
245.7	23\\
258.3	0\\
};
\addlegendentry{(CAV, RSU)}

\addplot[area legend, draw=none, fill=mycolor3, fill opacity=0.2, forget plot]
table[row sep=crcr] {%
x	y\\
0	0\\
6.3	1605\\
18.9	5274\\
31.5	5234\\
44.1	3188\\
56.7	2838\\
69.3	1180\\
81.9	605\\
94.5	319\\
107.1	264\\
119.7	142\\
132.3	67\\
144.9	98\\
157.5	415\\
170.1	311\\
182.7	186\\
195.3	118\\
207.9	91\\
220.5	83\\
233.1	58\\
245.7	23\\
258.3	0\\
258.3	0\\
245.7	0\\
233.1	0\\
220.5	0\\
207.9	0\\
195.3	0\\
182.7	0\\
170.1	0\\
157.5	0\\
144.9	0\\
132.3	0\\
119.7	0\\
107.1	0\\
94.5	0\\
81.9	0\\
69.3	0\\
56.7	0\\
44.1	0\\
31.5	0\\
18.9	0\\
6.3	0\\
0	0\\
}--cycle;

\addplot [color=mycolor2, line width=1.0pt, dashed]
  table[row sep=crcr]{%
0	0\\
6.39999999999964	2593\\
19.1999999999998	5669\\
32	3980\\
44.8000000000002	2734\\
57.6000000000004	2168\\
70.3999999999996	1793\\
83.1999999999998	781\\
96	345\\
108.8	284\\
121.6	194\\
134.4	167\\
147.2	169\\
160	425\\
172.8	256\\
185.6	206\\
198.4	132\\
211.2	99\\
224	36\\
236.8	27\\
249.6	41\\
262.4	0\\
};
\addlegendentry{(Truck, RSU)}

\addplot[area legend, draw=none, fill=mycolor2, fill opacity=0.2, forget plot]
table[row sep=crcr] {%
x	y\\
0	0\\
6.4	2593\\
19.2	5669\\
32	3980\\
44.8	2734\\
57.6	2168\\
70.4	1793\\
83.2	781\\
96	345\\
108.8	284\\
121.6	194\\
134.4	167\\
147.2	169\\
160	425\\
172.8	256\\
185.6	206\\
198.4	132\\
211.2	99\\
224	36\\
236.8	27\\
249.6	41\\
262.4	0\\
262.4	0\\
249.6	0\\
236.8	0\\
224	0\\
211.2	0\\
198.4	0\\
185.6	0\\
172.8	0\\
160	0\\
147.2	0\\
134.4	0\\
121.6	0\\
108.8	0\\
96	0\\
83.2	0\\
70.4	0\\
57.6	0\\
44.8	0\\
32	0\\
19.2	0\\
6.4	0\\
0	0\\
}--cycle;

\addplot [color=mycolor1, line width=1.0pt, dash dot]
  table[row sep=crcr]{%
0	0\\
10	3895\\
30	6932\\
50	4606\\
70	3274\\
90	1329\\
110	819\\
130	379\\
150	261\\
170	252\\
190	69\\
210	48\\
230	51\\
250	51\\
270	23\\
290	18\\
310	17\\
330	23\\
350	21\\
370	21\\
390	10\\
410	0\\
};
\addlegendentry{(Truck, CAV)}

\addplot[area legend, draw=none, fill=mycolor1, fill opacity=0.2, forget plot]
table[row sep=crcr] {%
x	y\\
0	0\\
10	3895\\
30	6932\\
50	4606\\
70	3274\\
90	1329\\
110	819\\
130	379\\
150	261\\
170	252\\
190	69\\
210	48\\
230	51\\
250	51\\
270	23\\
290	18\\
310	17\\
330	23\\
350	21\\
370	21\\
390	10\\
410	0\\
410	0\\
390	0\\
370	0\\
350	0\\
330	0\\
310	0\\
290	0\\
270	0\\
250	0\\
230	0\\
210	0\\
190	0\\
170	0\\
150	0\\
130	0\\
110	0\\
90	0\\
70	0\\
50	0\\
30	0\\
10	0\\
0	0\\
}--cycle;
\end{axis}
\end{tikzpicture}%
        \vspace{-20pt}
        \caption{}
        \label{fig:agent_dis}
    \end{subfigure}
    \hspace{-0pt}
    \begin{subfigure}[t]{0.23\textwidth}
        \centering
%
%
\definecolor{mycolor1}{rgb}{0.20000,0.36078,0.40392}%
\definecolor{mycolor2}{rgb}{0.56863,0.18431,0.33725}%
\begin{tikzpicture}[font=\scriptsize]

\begin{axis}[%
major tick length=1.5pt,
width=2.0in,
height=1.2in,
at={(0in,0in)},
xmin=-0.2,
xmax=11,
xlabel={Speed (m$\cdot$s$^{-1}$)},
ymin=0,
ymax=7400,
ylabel={Counts},
axis background/.style={fill=white},
xmajorgrids=false,
ymajorgrids=true,
grid style={dashed, opacity=0.6},
legend style={fill=none, nodes={scale=1.0},legend cell align=left, align=left, font=\tiny,inner sep=0pt,outer sep=0pt, row sep=-3pt},
xlabel style={yshift=0.11in},
ylabel style={yshift=-0.08in},
ytick distance=1500,
scaled y ticks=base 10:-3
]
\addplot [color=mycolor1, line width=1.0pt]
  table[row sep=crcr]{%
0	0\\
0.400000000000091	3653\\
1.19999999999982	1359\\
2	1516\\
2.80000000000018	1782\\
3.59999999999991	2286\\
4.40000000000009	3339\\
5.19999999999982	2465\\
6	2318\\
6.80000000000018	3381\\
7.59999999999991	0\\
};
\addlegendentry{Tractor}

\addplot[area legend, draw=none, fill=mycolor1, fill opacity=0.2, forget plot]
table[row sep=crcr] {%
x	y\\
0	0\\
0	0\\
0.4	3653\\
1.2	1359\\
2	1516\\
2.8	1782\\
3.6	2286\\
4.4	3339\\
5.2	2465\\
6	2318\\
6.8	3381\\
7.6	0\\
7.6	0\\
6.8	0\\
6	0\\
5.2	0\\
4.4	0\\
3.6	0\\
2.8	0\\
2	0\\
1.2	0\\
0.4	0\\
0	0\\
0	0\\
}--cycle;
\addplot [color=mycolor2, line width=1.0pt, dashed]
  table[row sep=crcr]{%
0	0\\
0.399999999999636	7148\\
1.19999999999982	1164\\
2	1147\\
2.80000000000018	1255\\
3.60000000000036	1264\\
4.39999999999964	1027\\
5.19999999999982	1247\\
6	3379\\
6.80000000000018	3507\\
7.60000000000036	331\\
8.39999999999964	203\\
9.19999999999982	58\\
10	30\\
10.8000000000002	37\\
11.6000000000004	55\\
12.3999999999996	29\\
13.1999999999998	51\\
14	22\\
14.8000000000002	19\\
15.6000000000004	19\\
16.3999999999996	35\\
17.1999999999998	23\\
18	27\\
18.8000000000002	18\\
19.6000000000004	4\\
20.3999999999996	0\\
};
\addlegendentry{CAV}

\addplot[area legend, draw=none, fill=mycolor2, fill opacity=0.2, forget plot]
table[row sep=crcr] {%
x	y\\
0	0\\
0.4	7148\\
1.2	1164\\
2	1147\\
2.8	1255\\
3.6	1264\\
4.4	1027\\
5.2	1247\\
6	3379\\
6.8	3507\\
7.6	331\\
8.4	203\\
9.2	58\\
10	30\\
10.8	37\\
11.6	55\\
12.4	29\\
13.2	51\\
14	22\\
14.8	19\\
15.6	19\\
16.4	35\\
17.2	23\\
18	27\\
18.8	18\\
19.6	4\\
20.4	0\\
20.4	0\\
19.6	0\\
18.8	0\\
18	0\\
17.2	0\\
16.4	0\\
15.6	0\\
14.8	0\\
14	0\\
13.2	0\\
12.4	0\\
11.6	0\\
10.8	0\\
10	0\\
9.2	0\\
8.4	0\\
7.6	0\\
6.8	0\\
6	0\\
5.2	0\\
4.4	0\\
3.6	0\\
2.8	0\\
2	0\\
1.2	0\\
0.4	0\\
0	0\\
}--cycle;
\end{axis}
\end{tikzpicture}%
        \vspace{-20pt}
        \caption{}
        \label{fig:agent_speed}
    \end{subfigure}
    \hspace{-0pt}
    \begin{subfigure}[t]{0.23\textwidth}
        \centering
%
%
\definecolor{mycolor1}{rgb}{0.20000,0.36078,0.40392}%
\begin{tikzpicture}[font=\scriptsize]
\begin{axis}[%
major tick length=1.5pt,
width=2.0in,
height=1.2in,
at={(0.4in,0.488in)},
unbounded coords=jump,
xmin=-41,
xmax=41,
xlabel={Articulation angle ($^{\circ}$)},
ymode=log,
ymin=1,
ymax=16000,
yminorticks=true,
ylabel={Counts},
axis background/.style={fill=white},
xmajorgrids=false,
ymajorgrids=true,
grid style={dashed, opacity=0.6},
xlabel style={yshift=0.11in},
ylabel style={yshift=-0.08in}
]
\addplot [color=mycolor1, line width=1.0pt, forget plot]
  table[row sep=crcr]{%
-68.4052009702597	2\\
nan	nan\\
-67.4052009702597	1\\
-66.9052009702597	1\\
nan	nan\\
-65.4052009702597	1\\
nan	nan\\
-64.4052009702597	1\\
nan	nan\\
-61.4052009702597	2\\
nan	nan\\
-57.4052009702597	1\\
-56.9052009702597	1\\
nan	nan\\
-54.4052009702597	36\\
-53.9052009702597	1\\
-53.4052009702597	2\\
-52.9052009702597	3\\
-52.4052009702597	1\\
-51.9052009702597	2\\
-51.4052009702597	1\\
-50.4052009702597	1\\
-49.9052009702597	3\\
-49.4052009702597	4\\
-48.9052009702597	4\\
-48.4052009702597	3\\
-47.9052009702597	5\\
-47.4052009702597	5\\
-46.9052009702597	10\\
-46.4052009702597	15\\
-45.9052009702597	9.00000000000001\\
-45.4052009702597	11.0000000000001\\
-44.9052009702597	8\\
-44.4052009702597	10\\
-43.9052009702597	6.99999999999989\\
-43.4052009702597	6\\
-42.9052009702597	6.99999999999989\\
-42.4052009702597	8\\
-41.9052009702597	6\\
-41.4052009702597	6.99999999999989\\
-40.9052009702597	6\\
-40.4052009702597	8\\
-39.9052009702597	10\\
-39.4052009702597	8\\
-38.9052009702597	4\\
-38.4052009702597	8\\
-37.9052009702597	9.00000000000001\\
-37.4052009702597	9.00000000000001\\
-36.9052009702597	6.99999999999989\\
-35.9052009702597	6.99999999999989\\
-35.4052009702597	12\\
-34.9052009702597	6.99999999999989\\
-34.4052009702597	9.00000000000001\\
-33.9052009702597	8\\
-33.4052009702597	8\\
-32.9052009702597	6.99999999999989\\
-32.4052009702597	11.0000000000001\\
-31.9052009702597	8\\
-31.4052009702597	5\\
-30.9052009702597	6.99999999999989\\
-30.4052009702597	10\\
-29.9052009702597	9.00000000000001\\
-29.4052009702597	8\\
-28.9052009702597	5\\
-28.4052009702597	11.0000000000001\\
-27.9052009702597	4\\
-27.4052009702597	6\\
-26.9052009702597	12\\
-26.4052009702597	12.9999999999999\\
-25.9052009702597	12\\
-25.4052009702597	3\\
-24.9052009702597	10\\
-24.4052009702597	9.00000000000001\\
-23.9052009702597	9.00000000000001\\
-23.4052009702597	11.0000000000001\\
-22.9052009702597	6\\
-22.4052009702597	11.0000000000001\\
-21.9052009702597	9.00000000000001\\
-21.4052009702597	10\\
-20.9052009702597	8\\
-20.4052009702597	10\\
-19.9052009702597	11.0000000000001\\
-19.4052009702597	11.0000000000001\\
-18.9052009702597	6\\
-18.4052009702597	15\\
-17.9052009702597	20\\
-17.4052009702597	10\\
-16.9052009702597	12.9999999999999\\
-16.4052009702597	13.9999999999998\\
-15.9052009702597	16\\
-15.4052009702597	16\\
-14.9052009702597	12.9999999999999\\
-14.4052009702597	20\\
-13.9052009702597	16.9999999999998\\
-13.4052009702597	155.999999999999\\
-12.9052009702597	12\\
-12.4052009702597	18.9999999999998\\
-11.9052009702597	12\\
-11.4052009702597	23\\
-10.9052009702597	25.9999999999998\\
-10.4052009702597	15\\
-9.90520097025967	23\\
-9.40520097025967	27\\
-8.90520097025967	32\\
-8.40520097025967	87.000000000001\\
-7.90520097025967	61.0000000000005\\
-7.40520097025967	75.9999999999994\\
-6.90520097025967	45\\
-6.40520097025967	62.0000000000008\\
-5.90520097025967	62.0000000000008\\
-5.40520097025967	50\\
-4.90520097025967	48\\
-4.40520097025967	48\\
-3.90520097025967	72.0000000000001\\
-3.40520097025967	122.999999999999\\
-2.90520097025967	177.000000000002\\
-2.40520097025967	145.000000000002\\
-1.90520097025967	318\\
-1.40520097025967	535.000000000002\\
-0.905200970259671	1300.99999999999\\
-0.405200970259671	2233.99999999998\\
0.0947990297403294	10311\\
0.594799029740329	1557\\
1.09479902974033	551.000000000002\\
1.59479902974033	398.000000000006\\
2.09479902974033	215.000000000001\\
2.59479902974033	245\\
3.09479902974033	138\\
3.59479902974033	87.000000000001\\
4.09479902974033	67.0000000000006\\
4.59479902974033	58.0000000000007\\
5.09479902974033	67.9999999999992\\
5.59479902974033	70.0000000000012\\
6.09479902974033	82.9999999999989\\
6.59479902974033	77.9999999999994\\
7.09479902974033	56.9999999999995\\
7.59479902974033	69.0000000000001\\
8.09479902974033	45\\
8.59479902974033	55.9999999999991\\
9.09479902974033	48\\
9.59479902974033	70.0000000000012\\
10.0947990297403	87.000000000001\\
10.5947990297403	75\\
11.0947990297403	86.0000000000005\\
11.5947990297403	40\\
12.0947990297403	33.0000000000003\\
12.5947990297403	46.9999999999993\\
13.0947990297403	16\\
13.5947990297403	18.9999999999998\\
14.0947990297403	22.0000000000002\\
14.5947990297403	18.9999999999998\\
15.0947990297403	31.0000000000004\\
15.5947990297403	27\\
16.0947990297403	32\\
16.5947990297403	18.9999999999998\\
17.0947990297403	20.9999999999997\\
17.5947990297403	12\\
18.0947990297403	18.9999999999998\\
18.5947990297403	18\\
19.0947990297403	16\\
19.5947990297403	12\\
20.0947990297403	23\\
20.5947990297403	27\\
21.0947990297403	27\\
21.5947990297403	22.0000000000002\\
22.0947990297403	25\\
22.5947990297403	43.0000000000002\\
23.0947990297403	23\\
23.5947990297403	15\\
24.0947990297403	18.9999999999998\\
24.5947990297403	9.00000000000001\\
25.0947990297403	25.9999999999998\\
25.5947990297403	16.9999999999998\\
26.0947990297403	24\\
26.5947990297403	9.00000000000001\\
27.0947990297403	12\\
27.5947990297403	12.9999999999999\\
28.0947990297403	20.9999999999997\\
28.5947990297403	23\\
29.0947990297403	89.0000000000012\\
29.5947990297403	13.9999999999998\\
30.0947990297403	3\\
30.5947990297403	10\\
31.0947990297403	11.0000000000001\\
31.5947990297403	9.00000000000001\\
32.0947990297403	8\\
32.5947990297403	10\\
33.0947990297403	8\\
33.5947990297403	10\\
34.0947990297403	12.9999999999999\\
34.5947990297403	6.99999999999989\\
35.0947990297403	6.99999999999989\\
35.5947990297403	10\\
36.0947990297403	6.99999999999989\\
36.5947990297403	11.0000000000001\\
37.0947990297403	3\\
37.5947990297403	4\\
38.5947990297403	4\\
39.0947990297403	5\\
39.5947990297403	2\\
40.0947990297403	6.99999999999989\\
40.5947990297403	5\\
41.0947990297403	1\\
41.5947990297403	3\\
42.0947990297403	2\\
42.5947990297403	3\\
43.0947990297403	5\\
43.5947990297403	3\\
44.0947990297403	4\\
44.5947990297403	5\\
45.0947990297403	2\\
45.5947990297403	2\\
};
\draw [->, thick] (-15,800) -- (-25,800) node[midway, above, font=\tiny] {Turn left};
\draw [->, thick] (15,800) -- (25,800) node[midway, above, font=\tiny] {Turn right};
\end{axis}

\end{tikzpicture}%
        \vspace{-20pt}
        \caption{}
        \label{fig:agent_angle}
    \end{subfigure}
    \vfill
    \raggedright
    \hspace{0pt}
    \begin{subfigure}[t]{0.23\textwidth}
        \centering
        \vspace{-0.2cm}
%
%
\definecolor{mycolor1}{rgb}{0.20000,0.36078,0.40392}%
\definecolor{mycolor2}{rgb}{0.70588,0.60392,0.40392}%
\definecolor{mycolor3}{rgb}{0.56863,0.18431,0.33725}%
\definecolor{mycolor4}{rgb}{0.70588,0.59216,0.83922}%
\definecolor{mycolor5}{rgb}{0.58824,0.67843,0.78431}%
\definecolor{mycolor6}{rgb}{0.54510,0.61569,0.51373}%
\definecolor{mycolor7}{rgb}{0.29804,0.32941,0.32941}%
\begin{tikzpicture}

\begin{axis}[%
major tick length=1.5pt,
width=2.0in,
height=1.2in,
at={(0.4in,0.488in)},
log origin=infty,
xmin=0.4,
xmax=7.6,
xtick={1,1.7,3,4,5,6,7},
xticklabels={{Car},{Pedestrian},{Van},{Truck},{Bus},{Bicycle},{Motorcycle}},
xticklabel style={rotate=45, yshift=0.08in},
ymode=log,
ymin=1000,
ymax=1100000,
yminorticks=false,
ylabel={Counts},
axis background/.style={fill=white},
grid style={dashed, opacity=0.6},
xmajorgrids=false,
ymajorgrids=true,
yminorgrids=false,
ylabel style={yshift=-0.08in},
]
\addplot[ybar, bar width=0.9, fill=mycolor1, fill opacity=0.5, draw=black, area legend] table[row sep=crcr] {%
1	672351\\
};
\addplot[ybar, bar width=0.9, fill=mycolor2, fill opacity=0.5, draw=black, area legend] table[row sep=crcr] {%
2	229041\\
};
\addplot[ybar, bar width=0.9, fill=mycolor3, fill opacity=0.5, draw=black, area legend] table[row sep=crcr] {%
3	95275\\
};
\addplot[ybar, bar width=0.9, fill=mycolor4, fill opacity=0.5, draw=black, area legend] table[row sep=crcr] {%
4	86702\\
};
\addplot[ybar, bar width=0.9, fill=mycolor5, fill opacity=0.5, draw=black, area legend] table[row sep=crcr] {%
5	42642\\
};
\addplot[ybar, bar width=0.9, fill=mycolor6, fill opacity=0.5, draw=black, area legend] table[row sep=crcr] {%
6	1988\\
};
\addplot[ybar, bar width=0.9, fill=mycolor7, fill opacity=0.5, draw=black, area legend] table[row sep=crcr] {%
7	1968\\
};
\end{axis}
\end{tikzpicture}%
        \vspace{-26pt}
        \caption{}
        \label{fig:type}
    \end{subfigure}
    \hspace{2pt}
    \begin{subfigure}[t]{0.23\textwidth}
        \centering
        \vspace{-0.2cm}
%
%
\definecolor{mycolor1}{rgb}{0.20000,0.36078,0.40392}%
\definecolor{mycolor2}{rgb}{0.70588,0.60392,0.40392}%
\definecolor{mycolor3}{rgb}{0.56863,0.18431,0.33725}%
\definecolor{mycolor4}{rgb}{0.58824,0.67843,0.78431}%
\begin{tikzpicture}

\begin{axis}[%
major tick length=1.5pt,
width=2.0in,
height=1.2in,
at={(0.4in,0.488in)},
xmin=-5,
xmax=65,
xtick={0,10,20,30,40,50,60},
xticklabels={{0-10},{10-20},{20-30},{30-40},{40-50},{50-60},{60-70}},
xticklabel style={yshift=0.06in,rotate=45},
xlabel={Co-visible objects},
ymin=0,
ymax=9500,
ylabel={Counts},
axis background/.style={fill=white},
xmajorgrids=false,
ymajorgrids=true,
grid style={dashed, opacity=0.6},
legend style={fill=none, nodes={scale=1.0},legend cell align=left, align=left, font=\tiny,inner sep=1pt,outer sep=0pt, row sep=-3pt},
legend image post style={scale=0.65},
xlabel style={yshift=0.08in},
ylabel style={yshift=-0.08in},
ytick distance=1500,
scaled y ticks=base 10:-3
]
\addplot[ybar, bar width=2, fill=mycolor1, fill opacity=0.5, draw=black, area legend,bar shift=-3] table[row sep=crcr] {%
0	2816\\
10	6932\\
20	7710\\
30	3920\\
40	531\\
50	164\\
60	26\\
};
\addplot[forget plot, color=white!15!black] table[row sep=crcr] {%
-4.90909090909091	0\\
64.9090909090909	0\\
};
\addlegendentry{(Tractor, Trailer)}

\addplot[ybar, bar width=2, fill=mycolor2, fill opacity=0.5, draw=black, area legend, bar shift=-1] table[row sep=crcr] {%
0	5969\\
10	9027\\
20	5788\\
30	1122\\
40	83\\
50	97\\
60	13\\
};
\addplot[forget plot, color=white!15!black] table[row sep=crcr] {%
-4.90909090909091	0\\
64.9090909090909	0\\
};
\addlegendentry{(Truck, CAV)}

\addplot[ybar, bar width=2, fill=mycolor3, fill opacity=0.5, draw=black, area legend, bar shift=1] table[row sep=crcr] {%
0	3265\\
10	7175\\
20	7005\\
30	2992\\
40	1404\\
50	258\\
60	0\\
};
\addplot[forget plot, color=white!15!black] table[row sep=crcr] {%
-4.90909090909091	0\\
64.9090909090909	0\\
};
\addlegendentry{(Truck, RSU)}

\addplot[ybar, bar width=2, fill=mycolor4, fill opacity=0.5, draw=black, area legend, bar shift=3] table[row sep=crcr] {%
0	7361\\
10	9104\\
20	4492\\
30	1092\\
40	50\\
50	0\\
60	0\\
};
\addplot[forget plot, color=white!15!black] table[row sep=crcr] {%
-4.90909090909091	0\\
64.9090909090909	0\\
};
\addlegendentry{(CAV, RSU)}

\end{axis}
\end{tikzpicture}%
        \vspace{-22pt}
        \caption{}
        \label{fig:co_object}
    \end{subfigure}
    \hspace{-0pt}
    \begin{subfigure}[t]{0.23\textwidth}
        \centering
        \vspace{-0.2cm}
%
%
\definecolor{mycolor1}{rgb}{0.20000,0.36078,0.40392}%
\definecolor{mycolor2}{rgb}{0.56863,0.18431,0.33725}%
\begin{tikzpicture}

\begin{axis}[%
major tick length=1.5pt,
width=2.0in,
height=1.2in,
at={(0.4in,0.788in)},
xmin=-3,
xmax=200,
xlabel={Distance (m)},
ymin=0,
ymax=154741.4,
ylabel={Counts},
axis background/.style={fill=white},
xmajorgrids=false,
ymajorgrids=true,
grid style={dashed, opacity=0.6},
legend style={fill=none, nodes={scale=1.0},legend cell align=left, align=left, font=\tiny,inner sep=0pt,outer sep=0pt, row sep=-3pt},
xlabel style={yshift=0.08in},
ylabel style={yshift=-0.08in},
]
\addplot [color=mycolor1, line width=1pt]
  table[row sep=crcr]{%
0	0\\
2.54999999998836	44569\\
7.64999999999418	78602\\
12.75	106313\\
17.8500000000058	123655\\
22.9500000000116	124979\\
28.0499999999884	126594\\
33.1499999999942	139819\\
38.25	141085\\
43.3500000000058	140531\\
48.4500000000116	133494\\
53.5499999999884	123834\\
58.6499999999942	109612\\
63.75	97749\\
68.8500000000058	99403\\
73.9500000000116	78552\\
79.0499999999884	61740\\
84.1499999999942	54174\\
89.25	43499\\
94.3500000000058	35316\\
99.4500000000116	30652\\
104.549999999988	25434\\
109.649999999994	18810\\
114.75	11447\\
119.850000000006	3423\\
124.950000000012	3\\
130.049999999988	0\\
150.450000000012	0\\
155.549999999988	3959\\
160.649999999994	10493\\
165.75	9780\\
170.850000000006	7828\\
175.950000000012	4934\\
181.049999999988	1429\\
186.149999999994	1488\\
191.25	721\\
196.350000000006	18\\
201.450000000012	1\\
206.549999999988	0\\
};
\addlegendentry{Vehicle}

\addplot[area legend, draw=none, fill=mycolor1, fill opacity=0.2, forget plot]
table[row sep=crcr] {%
x	y\\
0	0\\
2.55	44569\\
7.65	78602\\
12.75	106313\\
17.85	123655\\
22.95	124979\\
28.05	126594\\
33.15	139819\\
38.25	141085\\
43.35	140531\\
48.45	133494\\
53.55	123834\\
58.65	109612\\
63.75	97749\\
68.85	99403\\
73.95	78552\\
79.05	61740\\
84.15	54174\\
89.25	43499\\
94.35	35316\\
99.45	30652\\
104.55	25434\\
109.65	18810\\
114.75	11447\\
119.85	3423\\
124.95	3\\
130.05	0\\
135.15	0\\
140.25	0\\
145.35	0\\
150.45	0\\
155.55	3959\\
160.65	10493\\
165.75	9780\\
170.85	7828\\
175.95	4934\\
181.05	1429\\
186.15	1488\\
191.25	721\\
196.35	18\\
201.45	1\\
206.55	0\\
206.55	0\\
201.45	0\\
196.35	0\\
191.25	0\\
186.15	0\\
181.05	0\\
175.95	0\\
170.85	0\\
165.75	0\\
160.65	0\\
155.55	0\\
150.45	0\\
145.35	0\\
140.25	0\\
135.15	0\\
130.05	0\\
124.95	0\\
119.85	0\\
114.75	0\\
109.65	0\\
104.55	0\\
99.45	0\\
94.35	0\\
89.25	0\\
84.15	0\\
79.05	0\\
73.95	0\\
68.85	0\\
63.75	0\\
58.65	0\\
53.55	0\\
48.45	0\\
43.35	0\\
38.25	0\\
33.15	0\\
28.05	0\\
22.95	0\\
17.85	0\\
12.75	0\\
7.65	0\\
2.55	0\\
0	0\\
}--cycle;
\addplot [color=mycolor2, line width=1pt, dashed]
  table[row sep=crcr]{%
0	0\\
2.55000000000291	3439\\
7.65000000000146	24631\\
12.75	33082\\
17.8499999999985	43065\\
22.9499999999971	46810\\
28.0500000000029	45191\\
33.1500000000015	42729\\
38.25	39189\\
43.3499999999985	35871\\
48.4499999999971	34354\\
53.5500000000029	29992\\
58.6500000000015	24716\\
63.75	19905\\
68.8499999999985	15552\\
73.9499999999971	10271\\
79.0500000000029	8833\\
84.1500000000015	6263\\
89.25	4019\\
94.3499999999985	2733\\
99.4499999999971	1961\\
104.550000000003	1556\\
109.650000000001	1039\\
114.75	476\\
119.849999999999	123\\
124.949999999997	0\\
196.349999999999	0\\
201.449999999997	2\\
206.550000000003	0\\
};
\addlegendentry{Pedestrian}

\addplot[area legend, draw=none, fill=mycolor2, fill opacity=0.2, forget plot]
table[row sep=crcr] {%
x	y\\
0	0\\
2.55	3439\\
7.65	24631\\
12.75	33082\\
17.85	43065\\
22.95	46810\\
28.05	45191\\
33.15	42729\\
38.25	39189\\
43.35	35871\\
48.45	34354\\
53.55	29992\\
58.65	24716\\
63.75	19905\\
68.85	15552\\
73.95	10271\\
79.05	8833\\
84.15	6263\\
89.25	4019\\
94.35	2733\\
99.45	1961\\
104.55	1556\\
109.65	1039\\
114.75	476\\
119.85	123\\
124.95	0\\
130.05	0\\
135.15	0\\
140.25	0\\
145.35	0\\
150.45	0\\
155.55	0\\
160.65	0\\
165.75	0\\
170.85	0\\
175.95	0\\
181.05	0\\
186.15	0\\
191.25	0\\
196.35	0\\
201.45	2\\
206.55	0\\
206.55	0\\
201.45	0\\
196.35	0\\
191.25	0\\
186.15	0\\
181.05	0\\
175.95	0\\
170.85	0\\
165.75	0\\
160.65	0\\
155.55	0\\
150.45	0\\
145.35	0\\
140.25	0\\
135.15	0\\
130.05	0\\
124.95	0\\
119.85	0\\
114.75	0\\
109.65	0\\
104.55	0\\
99.45	0\\
94.35	0\\
89.25	0\\
84.15	0\\
79.05	0\\
73.95	0\\
68.85	0\\
63.75	0\\
58.65	0\\
53.55	0\\
48.45	0\\
43.35	0\\
38.25	0\\
33.15	0\\
28.05	0\\
22.95	0\\
17.85	0\\
12.75	0\\
7.65	0\\
2.55	0\\
0	0\\
}--cycle;
\end{axis}
\end{tikzpicture}%
        \vspace{-15pt}
        \caption{}
        \label{fig:object_dis}
    \end{subfigure}
    \hspace{-0pt}
    \begin{subfigure}[t]{0.23\textwidth}
        \centering
        \vspace{-0.2cm}
%
%
\definecolor{mycolor1}{rgb}{0.20000,0.36078,0.40392}%
\definecolor{mycolor2}{rgb}{0.56863,0.18431,0.33725}%
\begin{tikzpicture}

\begin{axis}[%
major tick length=1.5pt,
width=2.0in,
height=1.2in,
at={(0.4in,0.488in)},
xmin=0,
xmax=12,
xlabel={Speed (m$\cdot$s$^{-1}$)},
ymin=0,
ymax=662712.6,
ylabel={Counts},
axis background/.style={fill=white},
xmajorgrids=false,
ymajorgrids=true,
grid style={dashed, opacity=0.6},
legend style={fill=none, nodes={scale=1.0},legend cell align=left, align=left, font=\tiny,inner sep=0pt,outer sep=0pt, row sep=-3pt},
xlabel style={yshift=0.08in},
ylabel style={yshift=-0.03in},
]
\addplot [color=mycolor1, line width=1pt]
  table[row sep=crcr]{%
0	0\\
1.05000000004657	602466\\
3.15000000002328	74291\\
5.25	105784\\
7.34999999997672	97015\\
9.44999999995343	7913\\
11.5500000000466	2684\\
13.6500000000233	3642\\
15.75	1362\\
17.8499999999767	984\\
19.9499999999534	829\\
22.0500000000466	0\\
};
\addlegendentry{Vehicle}

\addplot[area legend, draw=none, fill=mycolor1, fill opacity=0.2, forget plot]
table[row sep=crcr] {%
x	y\\
0	0\\
1.05	602466\\
3.15	74291\\
5.25	105784\\
7.35	97015\\
9.45	7913\\
11.55	2684\\
13.65	3642\\
15.75	1362\\
17.85	984\\
19.95	829\\
22.05	0\\
22.05	0\\
19.95	0\\
17.85	0\\
15.75	0\\
13.65	0\\
11.55	0\\
9.45	0\\
7.35	0\\
5.25	0\\
3.15	0\\
1.05	0\\
0	0\\
}--cycle;
\addplot [color=mycolor2, line width=1pt, dashed]
  table[row sep=crcr]{%
0	0\\
0.399999999994179	38346\\
1.19999999999709	120528\\
2	70056\\
2.80000000000291	335\\
3.60000000000582	3294\\
4.39999999999418	420\\
5.19999999999709	3\\
6	15\\
6.80000000000291	0\\
};
\addlegendentry{Pedestrian}

\addplot[area legend, draw=none, fill=mycolor2, fill opacity=0.2, forget plot]
table[row sep=crcr] {%
x	y\\
0	0\\
0.4	38346\\
1.2	120528\\
2	70056\\
2.8	335\\
3.6	3294\\
4.4	420\\
5.2	3\\
6	15\\
6.8	0\\
6.8	0\\
6	0\\
5.2	0\\
4.4	0\\
3.6	0\\
2.8	0\\
2	0\\
1.2	0\\
0.4	0\\
0	0\\
}--cycle;
\end{axis}
\end{tikzpicture}%
        \vspace{-15pt}
        \caption{}
        \label{fig:object_speed}
    \end{subfigure}
    \caption{
    (a) Frame counts per scenario. (b) Distance statistics between agents. (c) Speed statistics of agents. (d) Articulation angle counts between the tractor and trailer. (e) Object category counts. (f) Statistics of co-visible objects among agents. (g) Distance statistics between objects and agents. (h) Speed statistics of objects.}
    \label{fig:combined}
\end{figure*}
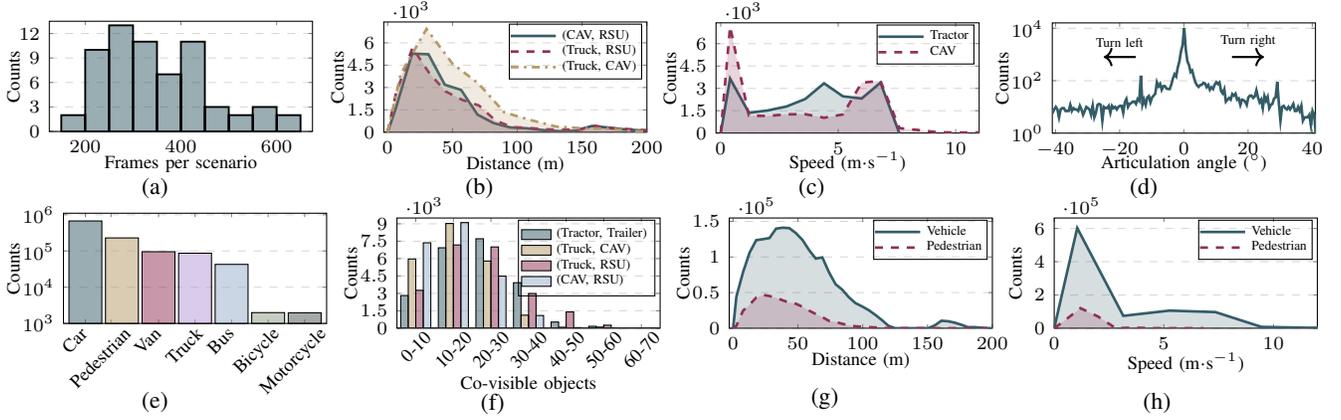

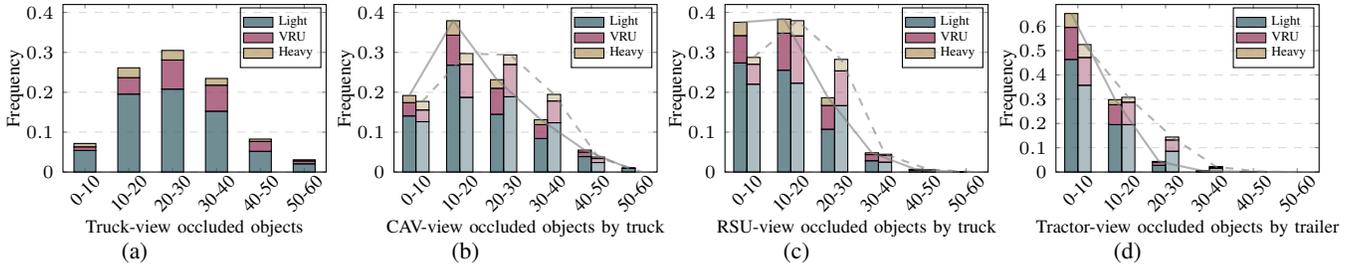
\begin{figure*}[htbp]
    \raggedright
    \hspace{-3pt}
    \begin{subfigure}[t]{0.2\textwidth}
        \centering
        \definecolor{mycolor1}{rgb}{0.20000,0.36078,0.40392}%
\definecolor{mycolor2}{rgb}{0.56863,0.18431,0.33725}%
\definecolor{mycolor3}{rgb}{0.70588,0.60392,0.40392}%
\begin{tikzpicture}

\begin{axis}[%
major tick length=1.5pt,
width=2.0in,
height=1.5in,
at={(0in,0in)},
bar width=2,
xmin=-5,
xmax=55,
xtick={0,10,20,30,40,50},
xticklabels={{0-10},{10-20},{20-30},{30-40},{40-50},{50-60}},
xticklabel style={yshift=0.06in,rotate=45},
xlabel={Truck-view occluded objects},
ymin=0,
ymax=0.42,
ylabel={Frequency},
axis background/.style={fill=white},
xmajorgrids=false,
ymajorgrids=true,
grid style={dashed, opacity=0.6},
legend style={fill=none, nodes={scale=1.0},legend cell align=left, align=left, font=\tiny,inner sep=1pt,outer sep=0pt, row sep=-3pt},
legend image post style={scale=0.65},
xlabel style={yshift=0.08in},
ylabel style={yshift=-0.08in},
ytick distance=0.1,
bar width=5
]
\addplot[ybar stacked,fill=mycolor1, fill opacity=0.7, draw=black, area legend] table[row sep=crcr] {%
0	0.053878420455352\\
10	0.19532566938829\\
20	0.207811673246949\\
30	0.152359289387687\\
40	0.0517688974527124\\
50	0.0204942015344619\\
};

\addlegendentry{Light}

\addplot[ybar stacked,fill=mycolor2, fill opacity=0.7, draw=black, area legend] table[row sep=crcr] {%
0	0.00921711649028491\\
10	0.0411762501502023\\
20	0.0729918886830708\\
30	0.0653737773247746\\
40	0.0249880672929379\\
50	0.00723904181934967\\
};

\addlegendentry{VRU}

\addplot[ybar stacked,fill=mycolor3, fill opacity=0.7, draw=black, area legend] table[row sep=crcr] {%
0	0.00849141269009324\\
10	0.0247768713570235\\
20	0.0240971123086335\\
30	0.0168929344640623\\
40	0.00582595755852637\\
50	0.00290162700231313\\
};
\addlegendentry{Heavy}

\end{axis}
\end{tikzpicture}%
        \vspace{-20pt}
        \caption{}
        \label{fig:occ_tt}
    \end{subfigure}
    \hspace{15pt}
    \begin{subfigure}[t]{0.2\textwidth}
        \centering
        \makeatletter
\newcommand\resetstackedplots{
\pgfplots@stacked@isfirstplottrue
}

\definecolor{mycolor1}{rgb}{0.20000,0.36078,0.40392}%
\definecolor{mycolor2}{rgb}{0.56863,0.18431,0.33725}%
\definecolor{mycolor3}{rgb}{0.70588,0.60392,0.40392}%
\begin{tikzpicture}

\begin{axis}[%
major tick length=1.5pt,
width=2.0in,
height=1.5in,
at={(0in,0in)},
bar width=2,
xmin=-5,
xmax=55,
xtick={0,10,20,30,40,50},
xticklabels={{0-10},{10-20},{20-30},{30-40},{40-50},{50-60}},
xticklabel style={yshift=0.06in,rotate=45},
xlabel={CAV-view occluded objects by truck},
ymin=0,
ymax=0.42,
ylabel={Frequency},
axis background/.style={fill=white},
xmajorgrids=false,
ymajorgrids=true,
grid style={dashed, opacity=0.6},
legend style={fill=none, nodes={scale=1.0},legend cell align=left, align=left, font=\tiny,inner sep=1pt,outer sep=0pt, row sep=-3pt},
legend image post style={scale=0.65},
xlabel style={yshift=0.08in},
ylabel style={yshift=-0.08in},
ytick distance=0.1
]
\addplot[ybar stacked, bar width=3, fill=mycolor1, fill opacity=0.7, draw=black, area legend, bar shift=-1.5] table[row sep=crcr] {%
0	0.140604528719977\\
10	0.267886424644781\\
20	0.144690364911893\\
30	0.0838820893583481\\
40	0.0386880353136808\\
50	0.00905025490763659\\
};

\addlegendentry{Light}

\addplot[ybar stacked, bar width=3, fill=mycolor2, fill opacity=0.7, draw=black, area legend, bar shift=-1.5] table[row sep=crcr] {%
0	0.0332857193999682\\
10	0.0752159997811975\\
20	0.0652365093746715\\
30	0.0349096572310418\\
40	0.0112147490468741\\
50	0.000633084628547686\\
};

\addlegendentry{VRU}

\addplot[ybar stacked, bar width=3, fill=mycolor3, fill opacity=0.7, draw=black, area legend, bar shift=-1.5] table[row sep=crcr] {%
0	0.0179427632558869\\
10	0.0361635124703515\\
20	0.0215605693708725\\
30	0.012191296982192\\
40	0.00509828883111706\\
50	8.2704679312606e-05\\
};

\addlegendentry{Heavy}

\resetstackedplots

\addplot[ybar stacked, bar width=3, fill=mycolor1, fill opacity=0.4, draw=black, area legend, bar shift=1.5, forget plot] table[row sep=crcr] {%
0	0.126142655613239\\
10	0.187013774422367\\
20	0.188634895955684\\
30	0.123533089877668\\
40	0.0238458469116904\\
50	0\\
};

\addplot[ybar stacked, bar width=3, fill=mycolor2, fill opacity=0.4, draw=black, area legend, bar shift=1.5, forget plot] table[row sep=crcr] {%
0	0.0294246523155316\\
10	0.0831127696874403\\
20	0.080961563817035\\
30	0.0545227689981075\\
40	0.0099048811438023\\
50	0\\
};

\addplot[ybar stacked, bar width=3, fill=mycolor3, fill opacity=0.4, draw=black, area legend, bar shift=1.5, forget plot] table[row sep=crcr] {%
0	0.0214468416525178\\
10	0.0270146629361061\\
20	0.0240795436924845\\
30	0.016573075574123\\
40	0.00378897740220296\\
50	0\\
};

\addplot [color=gray, line width=0.8pt, forget plot, opacity=0.6]
  table[row sep=crcr, x expr=\thisrowno{0}]{%
0-1.5	0.191833011375832\\
10-1.5	0.37926593689633\\
20-1.5	0.231487443657437\\
30-1.5	0.130983043571582\\
40-1.5	0.055001073191672\\
50-1.5	0.00976604421549689\\
};

\addplot [color=gray, dashed, line width=0.8pt, forget plot, opacity=0.6]
  table[row sep=crcr, x expr=\thisrowno{0}]{%
0+1.5	0.177014149581288\\
10+1.5	0.297141207045914\\
20+1.5	0.293676003465204\\
30+1.5	0.194628934449899\\
40+1.5	0.0375397054576956\\
50+1.5	0\\
};
\end{axis}
\end{tikzpicture}%
        \vspace{-20pt}
        \caption{}
        \label{fig:occ_cav}
    \end{subfigure}
    \hspace{15pt}
    \begin{subfigure}[t]{0.2\textwidth}
        \centering
        \makeatletter
\newcommand\resetstackedplots{
\pgfplots@stacked@isfirstplottrue
}

\definecolor{mycolor1}{rgb}{0.20000,0.36078,0.40392}%
\definecolor{mycolor2}{rgb}{0.56863,0.18431,0.33725}%
\definecolor{mycolor3}{rgb}{0.70588,0.60392,0.40392}%
\begin{tikzpicture}

\begin{axis}[%
major tick length=1.5pt,
width=2.0in,
height=1.5in,
at={(0in,0in)},
bar width=2,
xmin=-5,
xmax=55,
xtick={0,10,20,30,40,50},
xticklabels={{0-10},{10-20},{20-30},{30-40},{40-50},{50-60}},
xticklabel style={yshift=0.06in,rotate=45},
xlabel={RSU-view occluded objects by truck},
ymin=0,
ymax=0.42,
ylabel={Frequency},
axis background/.style={fill=white},
xmajorgrids=false,
ymajorgrids=true,
grid style={dashed, opacity=0.6},
legend style={fill=none, nodes={scale=1.0},legend cell align=left, align=left, font=\tiny,inner sep=1pt,outer sep=0pt, row sep=-3pt},
legend image post style={scale=0.65},
xlabel style={yshift=0.08in},
ylabel style={yshift=-0.08in},
ytick distance=0.1
]
\addplot[ybar stacked, bar width=3, fill=mycolor1, fill opacity=0.7, draw=black, area legend, bar shift=-1.5] table[row sep=crcr] {%
0	0.273466761153661\\
10	0.255273303933743\\
20	0.107135343127118\\
30	0.0282092809753847\\
40	0.00506265898821772\\
50	3.86349001931745e-05\\
};

\addlegendentry{Light}

\addplot[ybar stacked, bar width=3, fill=mycolor2, fill opacity=0.7, draw=black, area legend, bar shift=-1.5] table[row sep=crcr] {%
0	0.0685173798397462\\
10	0.092759282677192\\
20	0.0596488685844086\\
30	0.0155420817249479\\
40	0.00102962942919054\\
50	8.585533376261e-06\\
};

\addlegendentry{VRU}

\addplot[ybar stacked, bar width=3, fill=mycolor3, fill opacity=0.7, draw=black, area legend, bar shift=-1.5] table[row sep=crcr] {%
0	0.0334183058836048\\
10	0.0353651382227204\\
20	0.0195218625533373\\
30	0.00432762420672901\\
40	0.000668819116397275\\
50	6.43915003219575e-06\\
};

\addlegendentry{Heavy}

\resetstackedplots

\addplot[ybar stacked, bar width=3, fill=mycolor1, fill opacity=0.4, draw=black, area legend, bar shift=1.5, forget plot] table[row sep=crcr] {%
0	0.220125674765664\\
10	0.222730019675405\\
20	0.166403136930622\\
30	0.0245613308141938\\
40	0.0035665253228364\\
50	0\\
};

\addplot[ybar stacked, bar width=3, fill=mycolor2, fill opacity=0.4, draw=black, area legend, bar shift=1.5, forget plot] table[row sep=crcr] {%
0	0.0500854130858858\\
10	0.118498886139756\\
20	0.0871489764579644\\
30	0.0167698162071702\\
40	0.00240322507105187\\
50	0\\
};

\addplot[ybar stacked, bar width=3, fill=mycolor3, fill opacity=0.4, draw=black, area legend, bar shift=1.5, forget plot] table[row sep=crcr] {%
0	0.0171120423823515\\
10	0.0382108862726243\\
20	0.028861978439309\\
30	0.00313896559775237\\
40	0.000383122837414066\\
50	0\\
};

\addplot [color=gray, line width=0.8pt, forget plot, opacity=0.6]
  table[row sep=crcr, x expr=\thisrowno{0}]{%
0-1.5	0.375402446877012\\
10-1.5	0.383397724833655\\
20-1.5	0.186306074264864\\
30-1.5	0.0480789869070616\\
40-1.5	0.00676110753380554\\
50-1.5	5.36595836016313e-05\\
};

\addplot [color=gray, dashed, line width=0.8pt, forget plot, opacity=0.6]
  table[row sep=crcr, x expr=\thisrowno{0}]{%
0+1.5	0.287323130233901\\
10+1.5	0.379439792087785\\
20+1.5	0.282414091827895\\
30+1.5	0.0444701126191164\\
40+1.5	0.00635287323130234\\
50+1.5	0\\
};
\end{axis}
\end{tikzpicture}%
        \vspace{-20pt}
        \caption{}
        \label{fig:occ_rsu}
    \end{subfigure}
    \hspace{15pt}
    \begin{subfigure}[t]{0.2\textwidth}
        \centering
        \makeatletter
\newcommand\resetstackedplots{
\pgfplots@stacked@isfirstplottrue
}

\definecolor{mycolor1}{rgb}{0.20000,0.36078,0.40392}%
\definecolor{mycolor2}{rgb}{0.56863,0.18431,0.33725}%
\definecolor{mycolor3}{rgb}{0.70588,0.60392,0.40392}%
\begin{tikzpicture}

\begin{axis}[%
major tick length=1.5pt,
width=2.0in,
height=1.5in,
at={(0in,0in)},
bar width=2,
xmin=-5,
xmax=55,
xtick={0,10,20,30,40,50},
xticklabels={{0-10},{10-20},{20-30},{30-40},{40-50},{50-60}},
xticklabel style={yshift=0.06in,rotate=45},
xlabel={Tractor-view occluded objects by trailer},
ymin=0,
ymax=0.69,
ylabel={Frequency},
axis background/.style={fill=white},
xmajorgrids=false,
ymajorgrids=true,
grid style={dashed, opacity=0.6},
legend style={fill=none, nodes={scale=1.0},legend cell align=left, align=left, font=\tiny,inner sep=1pt,outer sep=0pt, row sep=-3pt},
legend image post style={scale=0.65},
xlabel style={yshift=0.08in},
ylabel style={yshift=-0.08in},
ytick distance=0.1
]
\addplot[ybar stacked, bar width=3, fill=mycolor1, fill opacity=0.7, draw=black, area legend, bar shift=-1.5] table[row sep=crcr] {%
0	0.463704787798439\\
10	0.195043411725278\\
20	0.0273423872627786\\
30	0.00247128109683129\\
40	0.000648177680421574\\
50	0\\
};

\addlegendentry{Light}

\addplot[ybar stacked, bar width=3, fill=mycolor2, fill opacity=0.7, draw=black, area legend, bar shift=-1.5] table[row sep=crcr] {%
0	0.132066110854984\\
10	0.0822265724177492\\
20	0.0132492432106674\\
30	0.00151665306601841\\
40	0.000371103662291655\\
50	0\\
};

\addlegendentry{VRU}

\addplot[ybar stacked, bar width=3, fill=mycolor3, fill opacity=0.7, draw=black, area legend, bar shift=-1.5] table[row sep=crcr] {%
0	0.0571589146112269\\
10	0.0206480240132291\\
20	0.00308727057828184\\
30	0.000412151692484058\\
40	5.39103293193959e-05\\
50	0\\
};

\addlegendentry{Heavy}

\resetstackedplots

\addplot[ybar stacked, bar width=3, fill=mycolor1, fill opacity=0.4, draw=black, area legend, bar shift=1.5, forget plot] table[row sep=crcr] {%
0	0.357258421514751\\
10	0.194999498414625\\
20	0.0853772206135459\\
30	0.0157462927417574\\
40	0.00113799789185891\\
50	0\\
};

\addplot[ybar stacked, bar width=3, fill=mycolor2, fill opacity=0.4, draw=black, area legend, bar shift=1.5, forget plot] table[row sep=crcr] {%
0	0.114422627310319\\
10	0.0925641491142246\\
20	0.046538235522968\\
30	0.00469670975522753\\
40	0.00027027449931649\\
50	0\\
};

\addplot[ybar stacked, bar width=3, fill=mycolor3, fill opacity=0.4, draw=black, area legend, bar shift=1.5, forget plot] table[row sep=crcr] {%
0	0.0532972936525503\\
10	0.020261937224255\\
20	0.0121792594280255\\
30	0.00121451988245486\\
40	3.55624341205908e-05\\
50	0\\
};

\addplot [color=gray, line width=0.8pt, forget plot, opacity=0.6]
  table[row sep=crcr, x expr=\thisrowno{0}]{%
0-1.5	0.652929813264649\\
10-1.5	0.297918008156257\\
20-1.5	0.0436789010517278\\
30-1.5	0.00440008585533376\\
40-1.5	0.00107319167203263\\
50-1.5	0\\
};

\addplot [color=gray, dashed, line width=0.8pt, forget plot, opacity=0.6]
  table[row sep=crcr, x expr=\thisrowno{0}]{%
0+1.5	0.524978342477621\\
10+1.5	0.307825584753104\\
20+1.5	0.144094715564539\\
30+1.5	0.0216575223794398\\
40+1.5	0.00144383482529599\\
50+1.5	0\\
};
\end{axis}
\end{tikzpicture}%
        \vspace{-20pt}
        \caption{}
        \label{fig:occ_t_t}
    \end{subfigure}

    \caption{Distribution of objects occluded for different agents in a 120m range. Distribution of objects occluded (a) from truck; (b) from CAV by truck; (c) from RSU by truck; (d) from the tractor by trailer. Solid lines linking left-side bars denote truck's straight driving scenarios, while dashed lines connecting right-side bars indicate truck's turning scenarios $(|\text{articulation angle}| > 5^\circ)$. Light vehicles, heavy vehicles and VRUs are represented by blue, yellow and red colors, respectively. }
    \label{fig:occ}
\end{figure*}

The dataset was built using CARLA's simulation platform, which provides an affordable way to recreate complex driving situations with realistic details. CARLA's flexible design allows the creation of different city environments with adjustable sensors, capturing high-quality data that includes precise pixel-level labels. In this system, semi-trucks, intelligent vehicles, and roadside devices work together through vehicle communication technology, forming a connected traffic network.
We generated scenarios by adding computer-controlled vehicles and pedestrians randomly appearing at key road locations. These digital traffic elements follow basic traffic rules when moving, but have some controlled randomness in their routes. The primary focus vehicles (semi-trucks and smart cars) are placed in areas with roadside sensors, positioned according to each map's layout and traffic patterns to cover critical urban driving situations.
Data collection included different environment types: country roads, busy city streets, highways, and various intersections (four-way, T-shaped, ramps, and roundabouts). The simulation used all available CARLA models, including 35 vehicle types (13 heavy vehicles and 22 light vehicles) plus 19 vulnerable road users (VRUs) types. CARLA's self-driving system and crash prevention features helped create natural traffic movements matching real-world behavior. Some typical scenarios are shown in Figure \ref{fig:occlusion_scene}.

\subsection{Dataset splits}
Our dataset follows the format of the OpenCOOD framework \cite{xu2022opv2v} for cooperative perception, comprising 64 representative scenarios with diverse vehicle and pedestrian types. Each scenario spans 17-60 seconds and includes multi-modal data from four agents:  tractor, trailer, CAV, and RSU, captured from multiple perspectives. Compared to existing datasets, our scenarios demonstrate more complex inter-agent occlusion patterns as illustrated in Figure \ref{fig:occlusion_scene}. Sensor data acquisition occurs at 10 Hz, yielding 88,396 synchronized LiDAR-image frames totaling 149.6 GB. 
The dataset is divided into train, validation, and test sets, with the training set containing 38 scenarios (13,243 frames), the validation set 9 scenarios (2,839 frames), and the testing set 17 scenarios (6,017 frames).

\subsection{Data statistics}
Figure \ref{fig:combined} provides some statistics in the dataset. The scenario duration analysis in Figure \ref{fig:scene} reveals an average of 345 frames per scenario, balancing the temporal continuity of cooperative behaviors with data efficiency. Spatial relationships between agents are quantified in Figure \ref{fig:agent_dis}, showing that truck-RSU and CAV-RSU distances predominantly remain under 80 meters, while truck-CAV interactions exhibit greater variability with occasional separations exceeding 80 meters.
Agent dynamics are further characterized in Figure \ref{fig:agent_speed}, where both truck and CAV primarily operate below 8 m$\cdot$s$^{-1}$. The CAV demonstrates occasional higher-speed maneuvers, while traffic light waiting phases explain observed low-speed instances. Cooperative perception analysis in Figure \ref{fig:co_object} indicates the tractor-trailer pair shares the most co-visible objects (average 53 per frame, peaking at 166), followed by truck-RSU and CAV-RSU pairs, with truck-CAV having the least overlap. Articulation patterns in Figure \ref{fig:agent_angle} reveal frequent straight-line motion alongside substantial turning/lane-change maneuvers.
Complementing these macroscopic insights, Figure \ref{fig:type} details object category distribution across seven classes, reflecting real-world traffic composition with light vehicles (65\%, cars and vans), VRUs (20\%, pedestrians, bicyclists, and motorcyclists), and heavy vehicles (15\%, truck and buses). Motion analysis in Figure \ref{fig:object_speed} shows most vehicles moving below 8 m$\cdot$s$^{-1}$ and pedestrian speeds clustering around $2$ m$\cdot$s$^{-1}$, while stationary vehicles predominantly originate from parking situations and traffic light queues. Proximity analysis in Figure \ref{fig:object_dis} demonstrates pedestrians are more likely than vehicles to appear near agents.

\begin{figure*}[h!]
    \centering
    \hspace{-10pt}
    \input{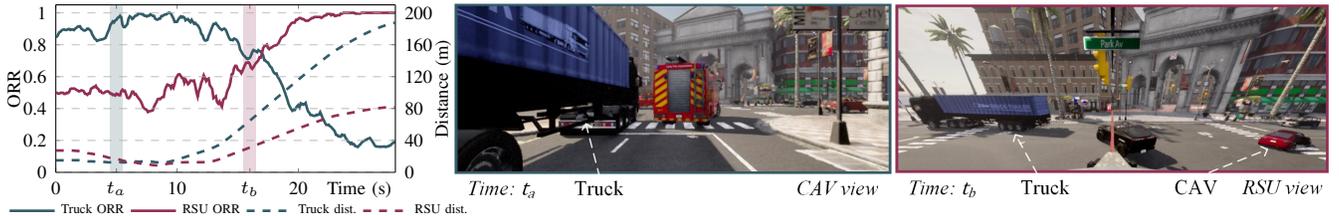}
    \caption{The temporal evolution of occlusion recovery rates (ORR) for CAV provided by truck and CAV, combined with the distances between truck, RSU, and CAV over time, in the  T-intersection scenario (left). At the critical time marker $t_a$ corresponding to CAV ingress into the intersection, substantial visual occlusion caused by the truck significantly impacts the CAV perception (center), manifesting as a high truck  occlusion recovery rate. As the CAV progresses through the intersection trajectory, by temporal milestone $t_b$ when approaching the egress zone, the occlusion from the truck diminishes significantly in the RSU's view (right), and the RSU occlusion recovery rate increases consequently.}
    \label{fig:occ_scene}
\end{figure*}

\section{Analysis of occlusions}
\subsection{Occlusion impacts}
TruckV2X captures extensive occlusion scenarios, which are systematically quantified in Figure \ref{fig:occ} through object counts across different agent perspectives. On the one hand, the number of occluded objects for the ego truck approximately follows a Gaussian distribution with a mean of 20-30 (Figure \ref{fig:occ_tt}). Due to the truck's higher viewpoint, these occlusions primarily originate from static environments like buildings.
On the other hand, it reveals that CAVs typically encounter 10-30 objects blocked by truck (Figure \ref{fig:occ_cav}), while elevated RSU viewpoints experience less severe impacts, with most cases falling within the 0-20 range (Figure \ref{fig:occ_rsu}). Notably, trucks exhibit intensified occlusion effects during turning maneuvers, producing measurable increases in obscured object counts. This phenomenon primarily stems from trailers generating localized occlusion patterns that specifically impact their corresponding tractors. While normal driving operations typically result in 0-10 occluded objects, turning conditions demonstrate obstruction levels rising sharply to 20-30 undetectable targets, as shown in Figure \ref{fig:occ_t_t}.

\subsection{Cooperation benefits}
Cooperative perception systems mitigate the impacts of occlusion by enabling agents to share perceptual information. To quantify the theoretical upper bound of cooperation benefits, we introduce the occlusion recovery rate, defined as the proportion of an ego agent’s occluded objects detected by cooperative ones to its total occluded objects, with values ranging from 0 to 1. An occlusion recovery rate close to 1 indicates nearly complete blind spot compensation, indicating optimal cooperative perception performance.

Figure \ref{fig:co_benefit} illustrates the distribution of occlusion recovery rates for various ego and cooperator configurations. For CAV, the cooperation benefits provided by truck are substantial in most scenarios, with the occlusion recovery rate exceeding 0.8 in over 60\% of cases, as shown in Figure \ref{fig:occ_cav_tt}. Moreover, most truck-CAV distances are within 50 meters in high occlusion recovery scenarios. As this distance increases, the cooperation benefits deteriorate rapidly.
In contrast, the cooperation benefits provided by RSU follow an entirely different trend (Figure \ref{fig:occ_cav_rsu}). The occlusion recovery rate predominantly falls between 0.4 and 0.6, and RSU is particularly effective in mitigating occlusions at near to mid ranges, whereas truck significantly reduces the occlusions it causes to CAV. 

For RSU, the cooperation benefits provided by truck exhibit a trend similar to that of truck benefiting CAV (Figure \ref{fig:occ_rsu_tt}), with an even higher proportion of cases where the occlusion recovery rate exceeds 0.8. However, the cooperative contribution of CAV to RSU is relatively weak (Figure \ref{fig:occ_rsu_cav}), with the occlusion recovery rate mostly below 0.4.
For truck, RSU provides slightly greater cooperation benefits than CAV, with occlusion recovery rates primarily between 0.4 and 0.6 (Figure \ref{fig:occ_tt_cav}), while those from CAV are concentrated between 0.2 and 0.4 (Figure \ref{fig:occ_tt_rsu}). Additionally, as the distance to truck decreases, the cooperation benefits from both RSU and CAV decline. 

Figure \ref{fig:occ_scene} illustrates the truck's occlusion effects through an intersection scenario. The plot shows inverse trends in occlusion recovery rates: truck's rate decreases over time while RSU's increases. This occurs because truck initially blocks the CAV's view when adjacent, making its sensor data crucial for compensating blind zones. As the CAV progresses through the intersection, truck's influence diminishes while the elevated RSU gradually becomes the dominant perception source. 

The above analysis reveals trucks' dual role as occlusion sources and potential mobile perception platforms, with their dynamic sensing enabling collaborative blind zone reduction for occlusion-resilient autonomous trucking.

\section{Benchmark experiments}
\begin{figure}[htbp]
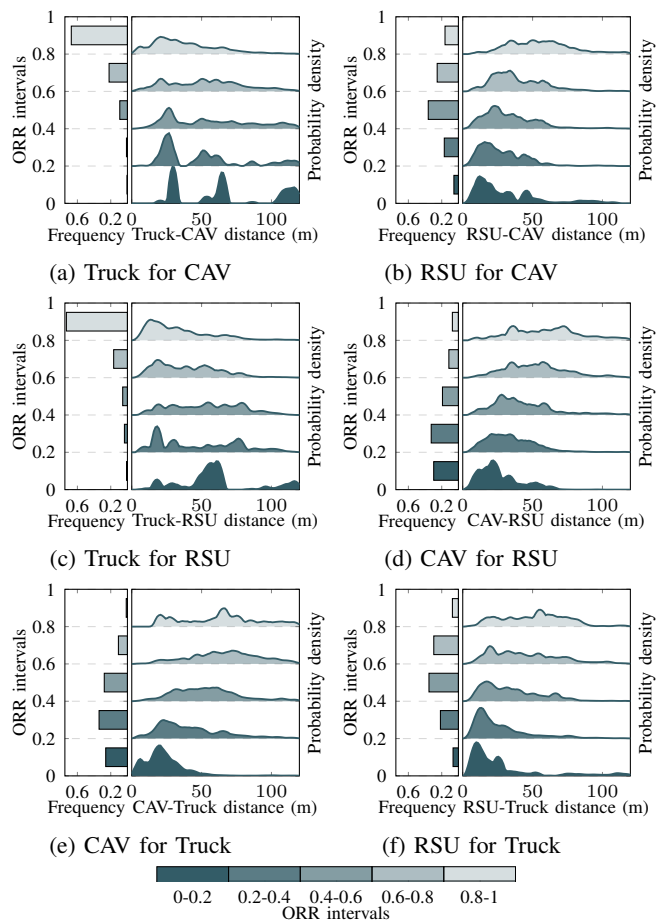

    \centering
    \hspace{-30pt}
    \begin{subfigure}[t]{0.2\textwidth}
        \centering
        \input{figures/occ_cav_tt}
        \vspace{-15pt}
        \caption{Truck for CAV}
        \label{fig:occ_cav_tt}
    \end{subfigure}
    \hspace{15pt}
    \begin{subfigure}[t]{0.2\textwidth}
        \centering
        \input{figures/occ_cav_rsu}
        \vspace{-15pt}
        \caption{RSU for CAV}
        \label{fig:occ_cav_rsu}
    \end{subfigure}
    \\
    \centering
    \hspace{-30pt}
    \begin{subfigure}[t]{0.2\textwidth}
        \centering
        \input{figures/occ_rsu_tt}
        \vspace{-15pt}
        \caption{Truck for RSU}
        \label{fig:occ_rsu_tt}
    \end{subfigure}
    \hspace{15pt}
    \begin{subfigure}[t]{0.2\textwidth}
        \centering
        \input{figures/occ_rsu_cav}
        \vspace{-15pt}
        \caption{CAV for RSU}
        \label{fig:occ_rsu_cav}
    \end{subfigure}
    \\
    \centering
    \hspace{-30pt}
    \begin{subfigure}[t]{0.2\textwidth}
        \centering
        \input{figures/occ_tt_cav}
        \vspace{-15pt}
        \caption{CAV for Truck}
        \label{fig:occ_tt_cav}
    \end{subfigure}
    \hspace{15pt}
    \begin{subfigure}[t]{0.2\textwidth}
        \centering
        \input{figures/occ_tt_rsu}
        \vspace{-15pt}
        \caption{RSU for Truck}
        \label{fig:occ_tt_rsu}
    \end{subfigure}
    \\[0.03in]
    \centering
    \hspace{-40pt}
    \begin{subfigure}[t]{0.2\textwidth}
        \centering
        \definecolor{mycolor1}{rgb}{0.20000,0.36078,0.40392}%
\begin{tikzpicture}
\begin{axis}[
    major tick length=1.5pt,
    axis x line=bottom,
    axis line style={-},
    axis y line=none,
    xtick=\empty,
    ytick=\empty,
    major tick length=0pt,
    xlabel style={align=center, yshift=0.2cm},
    xlabel={ORR intervals}, 
    xmin=0,
    xmax=1,
    ymin=0,
    ymax=0.2, 
    width=2.5in,   
    height=0.8in,  
    bar width=0.15in, 
    every axis plot/.append style={draw=black},
    xtick={0.1,0.3,0.5,0.7,0.9},
    xticklabels={{0-0.2},{0.2-0.4},{0.4-0.6},{0.6-0.8},{0.8-1}},
    at={(0in,0in)},
]

\addplot[xbar stacked, fill=mycolor1, fill opacity=1.0] table[row sep=crcr] {0.2 0\\};
\addplot[xbar stacked, fill=mycolor1, fill opacity=0.8] table[row sep=crcr] {0.2 0\\};
\addplot[xbar stacked, fill=mycolor1, fill opacity=0.6] table[row sep=crcr] {0.2 0\\};
\addplot[xbar stacked, fill=mycolor1, fill opacity=0.4] table[row sep=crcr] {0.2 0\\};
\addplot[xbar stacked, fill=mycolor1, fill opacity=0.2] table[row sep=crcr] {0.2 0\\};
\end{axis}
\end{tikzpicture}
        \vspace{-15pt}
    \end{subfigure}
    \caption{Frequency distribution of occlusion recovery rates (ORR) across various ego and cooperator configurations, with superimposed probability density distributions of inter-agent distances stratified by ORR intervals.}
    \label{fig:co_benefit}
\end{figure}

Following standard practices in truck perception \cite{fent2024man} and cooperative perception research \cite{xu2022opv2v,yu2022dair}, we establish object detection benchmarks using mean Average Precision (mAP) with Intersection-over-Union (IoU) thresholds of 0.3, 0.5, and 0.7 for three object categories: heavy vehicles, light vehicles, and VRUs.
The evaluation framework incorporates multiple ego agent configurations (as per Figure \ref{fig:co_benefit}'s cooperative benefit variations) within a detection range of $x \in [-128, 128]$m and $y \in [-64, 64]$m. Agent communication is constrained to a 300 m to reflect realistic vehicular networking capabilities. 
Due to significant occlusions in our TruckV2X dataset, we filter out bounding boxes for each agent where the number of LiDAR points within the box is below category-specific thresholds (5 points for VRUs and 15 for vehicles). 

\subsection{Experimental setup}

\begin{table*}[t]
\centering
\caption{Benchmark detection results of SOTA (state-of-the-art) cooperative perception methods on TruckV2X, with truck, CAV, and RSU acting as ego in mutual collaboration, where tractor and trailer are integrated via early fusion to function as a unified agent truck. All values are reported as percentages. Detection accuracy is evaluated using Average Precision (AP) and mean Average Precision (mAP) for light vehicles, heavy vehicles, and VRUs at IoU thresholds of 0.3, 0.5, and 0.7.}
\definecolor{color}{RGB}{255, 0, 0}   
\setlength{\tabcolsep}{2.2pt} 
\renewcommand\arraystretch{1.1} 

\begin{tabularx}{1.0\textwidth}
{>{\centering\arraybackslash}X|
c|
>{\centering\arraybackslash}X 
>{\centering\arraybackslash}X
>{\centering\arraybackslash}X|
>{\centering\arraybackslash}X
>{\centering\arraybackslash}X
>{\centering\arraybackslash}X|
>{\centering\arraybackslash}X
>{\centering\arraybackslash}X
>{\centering\arraybackslash}X|
>{\centering\arraybackslash}X
>{\centering\arraybackslash}X
>{\centering\arraybackslash}X
}
    \toprule
    \multirow{2}{*}{{Ego}} &
    \multirow{2}{*}{{Method}} & \multicolumn{3}{c|}{{$\text{AP}_{\text{light}}\text{@IoU}$}} & \multicolumn{3}{c|}{{$\text{AP}_{\text{heavy}}\text{@IoU}$}} & \multicolumn{3}{c|}{{$\text{AP}_{\text{VRU}}\text{@IoU}$}} & \multicolumn{3}{c}{{$\text{mAP}\text{@IoU}$}}
    \\
    \cline{3-14}
    &
    &{0.3}&{0.5}&{0.7}&{0.3}&{0.5}&{0.7}&{0.3}&{0.5}&{0.7}&{0.3}&{0.5}&{0.7}\\
    \hline  
    \multirow{8}{\hsize}{
    \parbox[c]{\hsize}{\centering {Truck (tractor and trailer)}}} & 
    Ego only 
    & 51.16 & 47.43 & 33.31
    & 55.27 & 53.99 & 48.62
    & 10.68 & 2.91 & 0.20
    & 39.03 & 34.78 & 27.38 \\
    \cline{2-14} 
    & Late Fusion 
    & 72.15 & 68.56 & 53.27
    & 53.41 & 52.56 & 48.02
    & 20.66 & 6.29 & 0.46
    & 48.74 & 42.47 & 33.92 \\
    \cline{2-14}
    & Early Fusion 
    & \textbf{77.22} & 63.25 & 49.17
    & \textbf{71.50} & \textbf{70.86} & \textbf{67.14}
    & \textbf{25.75} & 9.52 & 0.87
    & \textbf{58.16} & 47.88 & 39.06 \\
    \cline{2-14}
    & F-Cooper \cite{chen2019f}
    & 67.38 & 65.90 & 56.20
    & 65.77 & 65.61 & 62.57
    & 18.67 & 7.45 & 0.82
    & 50.61 & 46.65 & 39.86 \\
    & V2VNet \cite{wang2020v2vnet} 
    & 72.89 & \underline{71.55} & \underline{63.56}
    & \underline{68.90} & \underline{68.70} & 64.03
    & 19.74 & 8.52 & 0.95
    & 53.84 & 49.59 & \underline{42.85} \\ 
    & AttFuse \cite{xu2022opv2v} 
    & \underline{73.75} & \textbf{72.75} & \textbf{63.74}
    & 67.79 & 67.53 & \underline{64.79}
    & 22.24 & \underline{10.27} & \underline{1.16}
    & 54.59 & \textbf{50.18} & \textbf{43.23} \\
    & CoBEVT \cite{xu2022cobevt} 
    & 71.93 & 70.21 & 60.24
    & 67.82 & 67.35 & 63.94
    & \underline{25.53} & \textbf{11.91} & \textbf{1.35}
    & \underline{55.09} & \underline{49.83} & 41.84 \\
    & ERMVP \cite{zhang2024ermvp}
    & 67.93 & 66.33 & 56.53
    & 65.78 & 65.47 & 62.50
    & 21.63 & 9.37 & 0.94
    & 51.78 & 47.39 & 39.99 \\
    \hline
    \multirow{8}{*}{CAV} & 
    Ego only 
    & 48.13 & 46.01 & 35.99
    & 38.73 & 34.71 & 25.61
    & 10.35 & 3.39 & 0.28
    & 32.40 & 28.04 & 20.63 \\
    \cline{2-14} 
    & Late Fusion 
    & 72.73 & 70.46 & 58.34
    & 53.44 & 52.59 & 48.60
    & 22.86 & 7.89 & 0.64
    & 49.68 & 43.65 & 35.86 \\
    \cline{2-14}
    & Early Fusion 
    & \textbf{79.10} & \textbf{76.97} & \underline{67.08}
    & \textbf{71.96} & \textbf{70.89} & \textbf{65.79}
    & \textbf{29.65} & \textbf{11.68} & \textbf{1.04}
    & \textbf{60.24} & \textbf{53.18} & \underline{44.64} \\
    \cline{2-14}
    & F-Cooper \cite{chen2019f}
    & 75.12 & 73.85 & 65.01
    & 64.26 & 63.86 & 59.25
    & 18.32 & 6.39 & 0.50
    & 52.57 & 47.70 & 40.92 \\
    & V2VNet \cite{wang2020v2vnet} 
    & 75.35 & 74.28 & 65.70
    & 67.95 & 67.54 & 60.85
    & 18.09 & 6.37 & 0.55
    & 53.80 & 49.40 & 42.32 \\ 
    & AttFuse \cite{xu2022opv2v} 
    & \underline{77.54} & \underline{76.63} & \textbf{69.43}
    & \underline{69.91} & \underline{69.57} & \underline{64.26}
    & \underline{23.05} & \underline{9.02} & \underline{0.90}
    & \underline{56.83} & \underline{51.74} & \textbf{44.86} \\
    & CoBEVT \cite{xu2022cobevt} 
    & 74.62 & 72.98 & 62.39
    & 64.45 & 63.77 & 57.56
    & 20.67 & 8.56 & 0.85
    & 53.25 & 48.44 & 40.27 \\
    & ERMVP \cite{zhang2024ermvp}
    & 75.47 & 74.00 & 64.20
    & 61.97 & 61.46 & 55.45
    & 22.66 & 8.32 & 0.72
    & 53.36 & 47.93 & 39.79 \\
    \hline
    \multirow{8}{*}{RSU} & 
    Ego only 
    & 51.54 & 48.37 & 35.95
    & 50.45 & 47.99 & 41.10
    & 14.72 & 4.82 & 0.41
    & 38.90 & 33.73 & 25.82 \\
    \cline{2-14} 
    & Late Fusion 
    & \underline{73.50} & \underline{70.60} & \underline{57.37}
    & 49.13 & 48.18 & 43.71
    & 19.84 & 6.16 & 0.46
    & 47.49 & 41.65 & 33.85 \\
    \cline{2-14}
    & Early Fusion 
    & \textbf{76.82} & \textbf{73.44} & \textbf{59.77}
    & \textbf{70.06} & \textbf{68.42} & \textbf{63.51}
    & \textbf{30.12} & \textbf{11.20} & \underline{1.04}
    & \textbf{59.00} & \textbf{51.02} & \textbf{41.44} \\
    \cline{2-14}
    & F-Cooper \cite{chen2019f}
    & 60.75 & 60.12 & 53.58
    & 59.77 & 58.74 & 53.74
    & 19.77 & 8.21 & 0.92
    & 46.76 & 42.36 & 36.75 \\
    & V2VNet \cite{wang2020v2vnet} 
    & 65.60 & 64.65 & 56.07
    & \underline{63.22} & \underline{62.65} & 57.71
    & 21.20 & 8.38 & 0.82
    & 50.01 & 45.14 & 38.53 \\ 
    & AttFuse \cite{xu2022opv2v} 
    & 64.53 & 63.62 & 57.30
    & 62.77 & 62.39 & \underline{59.17}
    & \underline{26.03} & \underline{11.17} & \textbf{1.19}
    & \underline{51.11} & \underline{45.73} & \underline{39.22} \\
    & CoBEVT \cite{xu2022cobevt} 
    & 61.10 & 60.36 & 52.87
    & 51.68 & 51.07 & 46.26
    & 23.49 & 10.28 & 1.02
    & 45.42 & 40.57 & 33.38 \\
    & ERMVP \cite{zhang2024ermvp}
    & 63.39 & 62.50 & 55.34
    & 56.08 & 54.50 & 49.95
    & 20.52 & 8.24 & 0.81
    & 46.66 & 41.75 & 35.34 \\
    \bottomrule
\end{tabularx}
\label{tab:experiments}
\end{table*}

\begin{table*}[t]
\centering
\caption{Benchmark detection results of SOTA (state-of-the-art) cooperative perception methods on TruckV2X, with tractor (ego) and trailer collaborating. All values are reported as percentages. Detection accuracy is evaluated using Average Precision (AP) and mean Average Precision (mAP) for light vehicles, heavy vehicles, and VRUs at IoU thresholds of 0.3, 0.5, and 0.7.}
\definecolor{color}{RGB}{255, 0, 0}   
\setlength{\tabcolsep}{2.2pt} 
\renewcommand\arraystretch{1.1} 

\begin{tabularx}{1.0\textwidth}
{>{\centering\arraybackslash}X|
c|
>{\centering\arraybackslash}X 
>{\centering\arraybackslash}X
>{\centering\arraybackslash}X|
>{\centering\arraybackslash}X
>{\centering\arraybackslash}X
>{\centering\arraybackslash}X|
>{\centering\arraybackslash}X
>{\centering\arraybackslash}X
>{\centering\arraybackslash}X|
>{\centering\arraybackslash}X
>{\centering\arraybackslash}X
>{\centering\arraybackslash}X
}
    \toprule
    \multirow{2}{*}{{Ego}} &
    \multirow{2}{*}{{Method}} & \multicolumn{3}{c|}{{$\text{AP}_{\text{light}}\text{@IoU}$}} & \multicolumn{3}{c|}{{$\text{AP}_{\text{heavy}}\text{@IoU}$}} & \multicolumn{3}{c|}{{$\text{AP}_{\text{VRU}}\text{@IoU}$}} & \multicolumn{3}{c}{{$\text{mAP}\text{@IoU}$}}
    \\
    \cline{3-14}
    &
    &{0.3}&{0.5}&{0.7}&{0.3}&{0.5}&{0.7}&{0.3}&{0.5}&{0.7}&{0.3}&{0.5}&{0.7}\\
    \hline  
    \multirow{8}{*}{Tractor} & 
    Ego only 
    & 58.80 & 51.18 & 32.71
    & 62.63 & 61.26 & 56.28
    & 16.93 & 5.45 & 0.43
    & 46.12 & 39.30 & 29.81 \\
    \cline{2-14} 
    & Late Fusion 
    & \underline{71.43} & 61.83 & 42.15
    & 65.06 & 64.02 & 59.15
    & 24.11 & 8.32 & 0.71
    & 53.53 & 44.72 & 34.00 \\
    \cline{2-14}
    & Early Fusion 
    & \textbf{76.70} & 63.40 & 48.00
    & \textbf{73.06} & \textbf{72.19} & \textbf{67.66}
    & \textbf{28.51} & \textbf{11.88} & \underline{1.19}
    & \textbf{59.42} & \textbf{49.16} & 38.95 \\
    \cline{2-14}
    & F-Cooper \cite{chen2019f}
    & 66.45 & 65.25 & 55.28
    & 68.89 & 68.43 & 63.34
    & 19.97 & 8.36 & 0.85
    & 51.77 & 47.35 & 39.82 \\
    & V2VNet \cite{wang2020v2vnet} 
    & 71.00 & \textbf{69.84} & \textbf{60.72}
    & 68.13 & 67.59 & 63.75
    & 17.73 & 7.46 & 0.77
    & 52.29 & 48.30 & \textbf{41.75} \\ 
    & AttFuse \cite{xu2022opv2v} 
    & 65.94 & 64.91 & 55.97
    & \underline{69.76} & \underline{69.28} & \underline{64.83}
    & \underline{22.98} & \underline{10.16} & \textbf{1.20}
    & \underline{52.89} & \underline{48.12} & \underline{40.67} \\
    & CoBEVT \cite{xu2022cobevt} 
    & 66.54 & 65.07 & 53.88
    & 63.03 & 62.61 & 58.07
    & 19.85 & 8.70 & 0.91
    & 49.81 & 45.46 & 37.62 \\
    & ERMVP \cite{zhang2024ermvp}
    & 69.56 & \underline{67.92} & \underline{57.40}
    & 62.90 & 62.40 & 59.18
    & 19.36 & 7.52 & 0.69
    & 50.61 & 45.95 & 39.09 \\
    \bottomrule
\end{tabularx}

\label{tab:experiments_tt}
\end{table*}

To comprehensively test the cooperative perception baselines on TruckV2X, we build several models upon the OpenCOOD framework \cite{xu2022opv2v} and perform a comprehensive performance analysis.  
Specifically, we evaluate three cross-agent fusion strategies: early, intermediate, and late fusion.
In early fusion, LiDAR point clouds from collaborating agents are spatially aligned to the ego vehicle's coordinate system using shared pose data, creating consolidated point cloud inputs for detection models.
For intermediate fusion, we implement five state-of-the-art collaborative perception frameworks: F-Cooper \cite{chen2019f}, V2VNet \cite{wang2020v2vnet}, AttFuse \cite{xu2022opv2v}, CoBEVT \cite{xu2022cobevt}, and ERMVP \cite{zhang2024ermvp}. These methods enable multi-agent collaboration by fusing intermediate feature representations from individual perception models.
Late fusion adopts a decentralized paradigm where each agent shares its calibrated pose and confidence-scored detection boxes. The multi-source detections are geometrically projected onto the ego coordinate system and consolidated through standard non-maximum suppression (NMS).
All models are trained by 120 epochs with an initial learning rate of 0.001 using the Adam optimizer (weight decay $\lambda = 10^{-4}$). The learning policy combines cosine annealing scheduling with 20 warmup epochs. Three standard  point cloud augmentations are used: random scaling, rotation, and flipping. 

\subsection{Results and analysis}
The results are presented in Table \ref{tab:experiments} and \ref{tab:experiments_tt}.
For all categories, early fusion integrates point clouds with minimal information loss, resulting in higher AP than other methods.
For light vehicles and VRUs, the Average Precision  progressively improves from no fusion (ego-only baseline) through late fusion to early fusion configurations, regardless of the selected ego agent. Late fusion achieves substantially higher AP values than the no-fusion baseline, while intermediate fusion exhibits minimal performance variations.
In light vehicle detection scenarios, intermediate fusion approaches achieve early fusion performance levels when the CAV serves as the ego agent. When truck or tractor units assume the ego role, intermediate fusion performance matches late fusion baselines. Notably, RSU-based ego configurations show degraded intermediate fusion performance with over 10\% AP reduction compared to late fusion. This can be attributed to increasing discrepancies among CAV, truck, and RSU. Both truck and RSU have viewpoint differences relative to CAV, and RSU is static. As most intermediate fusion methods are CAV-centered, larger discrepancies lead to more pronounced performance degradation.
VRU detection displays distinct characteristics, with most intermediate fusion methods marginally surpassing late fusion performance. A subset of these methods even exceeds early fusion results. Heavy vehicle detection reveals contrasting behavior: late fusion underperforms the no-fusion baseline when truck or RSU units act as ego agents. Early fusion consistently outperforms late fusion by over 10\% AP across all configurations, while maintaining narrower performance gaps with intermediate fusion approaches.
All methods show minimal VRU detection differences, likely from smaller size and sparser point clouds worsening difficulty. Heavy vehicles’ large scale often shows only one side per viewpoint, causing late fusion degradation from low accuracy. Early and intermediate fusion aggregates multi-view info for full contours, boosting accuracy significantly.

Additionally, we observe that as the IoU threshold increases, early fusion AP for light vehicles drops significantly, especially when the ego is a truck or tractor. At an IoU of 0.7, early fusion underperforms all intermediate fusion methods. However, it is not an issue for heavy vehicles, where early fusion consistently achieves the highest AP. 
This may stem from relative pose errors between the tractor and trailer, causing incomplete point cloud alignment. Light vehicles' smaller size makes their AP more susceptible to such effects than heavy vehicles.
Overall, intermediate fusion performs comparably to early fusion when the ego is CAV but is often inferior to late fusion otherwise. This might stem from viewpoint differences between truck, RSU, and CAV. Further research is needed to explore cooperative perception involving truck and RSU, particularly interactions between the tractor and trailer.

\section{Conclusions}
We present TruckV2X, the first truck-centric cooperative perception dataset capturing complex, occluded scenarios with heavy vehicle interactions. 
It features synchronized multi-agent observations from tractors, trailers, CAVs, and RSUs, all equipped with comprehensive sensors. Our benchmarks establish baselines for object detection using state-of-the-art cooperative perception methods. Analysis highlights trucks' dual role as both occlusion sources and potential mobile perception platforms, with their dynamic sensing reducing blind spots for resilient autonomous trucking.
However, our dataset lacks diverse weather, pedestrian-vehicle hazards, and sensor noise. Beyond blind spots and occlusions, bridging these sim-to-real gaps demands continuous refinement in the real world. We hope this research stimulates new directions in truck-oriented cooperative perception and encourages practical contributions for deployment.

{\small
\bibliographystyle{IEEEtran}
\bibliography{references}
}

\end{document}